\PassOptionsToPackage{table}{xcolor}
\documentclass[11pt]{article}

\usepackage[preprint]{acl}

\usepackage{times}
\usepackage{latexsym}
\usepackage[T1]{fontenc}
\usepackage{hyperref}
\usepackage{url}
\usepackage{booktabs}
\usepackage{amsfonts}
\usepackage{amsmath}
\usepackage{amssymb}
\usepackage{mathtools}
\usepackage{amsthm}
\usepackage{nicefrac}
\usepackage{microtype}
\usepackage{xcolor}
\usepackage{graphicx}
\usepackage{subcaption}
\usepackage{multirow}
\usepackage{longtable}
\usepackage{caption}
\usepackage{enumitem}
\usepackage{tikz}
\usepackage{pifont}
\usepackage{algorithm}
\usepackage{algpseudocode}
\usetikzlibrary{positioning, fit, backgrounds, calc}
\usepackage{placeins}
\usepackage[capitalize,noabbrev]{cleveref}

\theoremstyle{plain}

\theoremstyle{definition}

\theoremstyle{remark}

\usepackage[T1]{fontenc}

\usepackage[utf8]{inputenc}

\usepackage{microtype}

\usepackage{inconsolata}

\usepackage{graphicx}

\title{Large Language Models Can Follow Instructions, But Not Many at Once: \\ Phase Transitions in Compositional Constraint Satisfaction}

  \author{Mariya I. Vasileva\thanks{Work done while at Meta Superintelligence Labs.} \\
    \texttt{maria.i.vasileva@gmail.com}}

\begin{document}
\maketitle

% ============================================================
% ============================================================
% ============================================================
% ============================================================

\begin{abstract}
Large language models are increasingly deployed in settings that require simultaneous adherence to multiple explicit constraints---reasoning structure, safety boundaries, output schemas.
Individual constraints are handled proficiently, but the compositional regime, where many must hold jointly, remains poorly characterized: how rapidly does performance degrade, what governs the degradation, and can the collapse be mitigated?
We introduce \textbf{Constraint Saturation Evaluation~(CSE)}, a procedurally generated benchmark that systematically varies the number of simultaneous constraints~($k$), with every constraint scored by a deterministic, rule-based verifier and zero LLM-judge involvement: 15 models, 36 constraint types, 369{,}753 checks at $k{=}1{-}12$.
Three findings emerge.
First, per-constraint pass rate decays gradually and predictably, while the chance of satisfying \emph{all} $k$ constraints collapses---a model passing individual constraints at ${\sim}41\%$ at $k{=}8$ succeeds on all eight just 5.7\% of the time.
Second, constraints do not degrade equally: structural constraints lose $2{\times}$ more baseline capability per added constraint than lexical ones, ordered by a comprehension-maintenance gap that separates constraints requiring sustained tracking from binary decisions immune to composition.
Third, failures are nearly independent, which is what makes the accumulation multiplicative; the residual coupling that does exist tracks shared output features rather than pairwise interference---a wrong sentence count fails every constraint that reads it.
Weak coupling means there is no pairing to exploit: no selection or arrangement of constraints mitigates the collapse.
Three targeted ablations find little room at inference time: pre-generation planning does not move the threshold at all, while post-hoc self-correction and best-of-5 retries delay it by only one to two constraints.
Only raising the per-constraint pass rate helps.
Reliable instruction following breaks down beyond 5--6 simultaneous constraints: probe-level success falls below 50\% at 7 constraints for the strongest model, and at 3 or fewer for 12 of 15.
All verifiers, probes, model outputs, and code are released for reproducibility.
\end{abstract}

\section{Introduction}
\label{sec:intro}

Language models deployed in complex applications must now adhere to multiple explicit constraints simultaneously---reasoning structure, safety boundaries, domain-specific protocols, output schemas---within a single response. Models handle individual constraints well, but performance under composition is poorly understood: does it degrade gracefully, or is there a sharp transition? Three specific questions remain open: what is the functional form of the degradation, is there a structure governing which constraints fail first, and does the collapse arise from pairwise interference between constraints or from something simpler? A fourth question follows from any answer to these: can the collapse be mitigated?

Several benchmarks address pieces of this question.
IFEval \cite{zhou2023instruction} pioneered deterministic verification across 25 constraint types but evaluates at $k{\leq}3$.
COLLIE \cite{yao2023collie} demonstrated through a formal constraint grammar that composition increases difficulty, but uses fixed compositions rather than varying $k$.
FollowBench \cite{jiang2024followbench} extends to $k{=}5$---the closest existing per-$k$ analysis---but relies on GPT-4 as judge.
RECAST \cite{guo2025recast} pushes to 13+ constraints but reports at four coarse tiers; CCTU \cite{cctu2025} finds no model exceeds 20\% under strict multi-constraint adherence.
InfoBench \cite{qin2024infobench} observes that models fail number and linguistic constraints preferentially---a hierarchy we extend across $k$.
Outside text, ConceptMix \cite{conceptmix2024} reports that text-to-image models drop from 83\% to 8\% as concept count grows from 1 to 7.
In compositional generalization more broadly \cite{lake2018generalization}, \citet{dziri2023faith} show exponential decay with composition depth, and \citet{press2023measuring} quantify a compositionality gap persisting at ${\sim}40\%$ regardless of scale.
Benchmarks that rely on LLM-based evaluation \cite{jiang2024followbench,guo2025recast,conceptmix2024} face a circularity at high $k$: the judge itself struggles with compositional constraint violations in exactly the regime where models fail most.
The pattern is otherwise consistent across benchmarks, modalities, and tasks: composition is hard.
What is missing is the functional form---where capability breaks, what drives the breakdown, and whether the trajectory is predictable.
An extended discussion of all related work, including training methods, mechanistic explanations, cross-modal parallels, and phase transition theory, appears in \cref{app:related_full}.

We introduce \textbf{Constraint Saturation Evaluation~(CSE)}, designed to address all three. Every constraint is verified by a deterministic function rather than an LLM judge---a design choice that makes CSE's measurements invariant to the same compositional degradation it studies. CSE evaluates 36 deterministically verifiable constraint types spanning 8 processing dimensions, 4{,}527 probes at $k{=}1$ to $12$, 15 models across 8 families (369{,}753 constraint checks).

We measure the co-failure structure directly and find failures nearly independent, which makes joint success approximately the product of $k$ per-constraint rates: a mild per-constraint decay compounds into a steep collapse. Prior work assumes interference. The rates, their 2\%-to-28\% spread across models, and the failure ordering are all experimental results.

We address three open questions:
\begin{enumerate}[leftmargin=*, itemsep=2pt, topsep=2pt]
\item \textbf{Is the decay predictable?} Per-constraint pass rate follows a multiplicative model ($72.0\% \times 0.922^{k-1}$, held-out MAE${=}0.2$pp), but probe-level success collapses below 2\% by $k{=}9$: a model passing individual constraints at ${\sim}41\%$ at $k{=}8$ succeeds on all eight just 5.7\% of the time. Probe-level success drops below 50\% at just 7 constraints for the strongest model, and at 3 or fewer for 12 of 15 (\cref{sec:experiments}).
\item \textbf{Do all constraints degrade equally?} Structural constraints lose $2.0{\times}$ more baseline capability per added constraint than lexical ones (95\% CI: $[1.9, 2.3]$), and the comprehension-maintenance gap ($\rho{=}{-}0.584$) predicts degradation rate: constraints requiring sustained tracking during generation degrade fastest, while binary decisions are immune (\cref{sec:degradation}).
\item \textbf{What drives the collapse---interference or accumulation?} Constraint failures are weakly correlated through shared output features (mean $\varphi{=}{+}0.067$), not pairwise interference. This near-multiplicative structure is what makes the decay in (1) predictable---and it means the only lever for improvement is per-constraint reliability, not constraint pairing (\cref{sec:cofailure}).
\end{enumerate}

\section{Related Work}
\label{sec:related}

Several benchmarks address pieces of this question.
IFEval \cite{zhou2023instruction} and COLLIE \cite{yao2023collie} establish deterministic verification of instruction-following constraints, with COLLIE demonstrating through a formal constraint grammar that composition increases difficulty---but IFEval evaluates at $k \leq 3$, and COLLIE uses fixed compositions rather than systematically varying $k$, precluding trajectory analysis.
FollowBench \cite{jiang2024followbench} introduces a multi-level mechanism that incrementally adds one constraint per level up to $k{=}5$ and tracks both hard and soft satisfaction rates across levels---the closest existing per-$k$ analysis---but relies on GPT-4 as judge for open-ended constraints and does not characterize the functional form of the observed degradation.
RECAST \cite{guo2025recast} pushes constraint density to 13+ per instance and proposes reinforcement learning from verifiable constraints, but reports performance at four coarse difficulty tiers rather than per-$k$ trajectories, precluding decay analysis.
In constrained tool use, CCTU \cite{cctu2025} finds no model exceeds 20\% task completion under strict multi-constraint adherence.
Outside text entirely, ConceptMix \cite{conceptmix2024} reports that text-to-image models drop from 83\% to 8\% as concept count grows from 1 to 7.
The pattern is consistent across benchmarks and modalities: composition is hard.
What is missing is a characterization of the \emph{functional form} of the degradation---where capability breaks, what drives the breakdown, and whether the trajectory is predictable.
An extended discussion of all related work, including training methods, mechanistic explanations, cross-modal parallels, and phase transition theory, appears in \cref{app:related_full}.

\section{Constraint Saturation Evaluation (CSE)}
\label{sec:method}

\subsection{Constraint taxonomy}
\label{sec:taxonomy}

CSE defines 36 constraint types across 8 dimensions (\cref{tab:constraint_profiles}), selected to be deterministically verifiable and to span a range of processing demands.
Constraints range from local pattern matching (lipogram: avoid a letter; forbidden word: exclude a token) through document-level structure (exact paragraph count, monotonic sentence length) to relational reasoning requiring structured output (scene graphs, logic grids) and self-referential reasoning (a sentence correctly reporting its own word count).
Whether this design choice produces a measurable hierarchy in degradation rates is tested in \cref{sec:degradation}. All 36 verifiers are fully deterministic, each returning both a binary pass/fail judgment and a continuous score $s \in [0,1]$ representing partial compliance.

Each constraint entered the pool through a four-stage vetting pipeline---deterministic verifiability, parameter calibration, composability analysis, and pilot validation (\cref{app:constraint_design}).
The composability stage is critical for a compositional benchmark: a constraint that individually tests an interesting capability but locks down multiple output dimensions (sentence count \emph{and} word count \emph{and} ordering) inherits the union of those dimensions' incompatibilities and would distort the composition bias at high~$k$.
Composability profiles for all 38~candidates and a case study comparing an accepted constraint against a rejected one appear in \cref{app:constraint_design}.

Following \citet{jiang2024followbench}, we report two complementary satisfaction metrics. For a set of $m$~probes, each containing $k$~constraints with binary outcomes $s^j_i \in \{0,1\}$:

Strict CSR (sCSR):
\begin{equation}
\text{sCSR} = \frac{1}{m} \sum_{i=1}^{m} \prod_{j=1}^{k} s^j_i
\label{eq:scsr}
\end{equation}
measures the fraction of probes where \emph{all} $k$~constraints are simultaneously satisfied.

Marginal CSR (mCSR):
\begin{equation}
\text{mCSR} = \frac{1}{mk} \sum_{i=1}^{m} \sum_{j=1}^{k} s^j_i
\label{eq:mcsr}
\end{equation}
measures the average per-constraint satisfaction rate.

At $k{=}1$, sCSR and mCSR are identical. Their divergence at higher $k$ is the phase transition: mCSR decays gradually (each constraint fails somewhat more often under load), but sCSR collapses because it is the \emph{product} of $k$~marginal rates.
sCSR is the deployment-relevant metric---in practice, all constraints must hold simultaneously---and is where the phase transition is visible.
mCSR shows that per-constraint competence is not the bottleneck: models that pass individual constraints at 45\% may succeed on complete probes only 12\% of the time.

We additionally report the mean continuous score $\bar{s}_i \in [0,1]$ for each constraint, which credits near-misses (e.g., a model producing 48~words when the target is 50 scores ${\sim}0.96$ but passes${=}$False).
The \emph{comprehension-maintenance gap} is
\begin{equation}
\Delta_i = \bar{s}_i - \text{mCSR}_i,
\label{eq:gap}
\end{equation}
where a large $\Delta$ indicates that a model partially complies---demonstrating comprehension of the constraint---but fails to maintain full satisfaction.

We define the \emph{compositional half-life} $k^*$ for each model as the smallest $k$ at which sCSR drops below 50\%---the number of simultaneous constraints a model can reliably handle.

\subsection{Probe generation}
\label{sec:probes}

CSE probes are generated procedurally (\cref{alg:probe_gen}).
For each constraint count $k$, the composer draws $k$-subsets from the constraint pool, rejects unsatisfiable combinations via the compatibility checker (\cref{app:compatibility}), instantiates parameters from fixed calibrated ranges, shuffles constraint presentation order, and assembles the prompt.
The composer targets a fixed number of satisfiable probes per $k$ level, resampling new combinations when rejections occur rather than accepting reduced yield.
At low $k$ (${\leq}5$), all $\binom{|\mathcal{C}|}{k}$ subsets are materialized and shuffled for uniform coverage; at $k{\geq}6$, combinations are sampled uniformly at random with a per-combination reuse cap.

\paragraph{Compatibility checker.}
The compatibility checker enforces satisfiability through unconditional pair blocks, parameter-dependent arithmetic checks, and cross-constraint propagation (\cref{app:compatibility}).
Rejection rates increase with $k$: from ${\sim}32\%$ at $k{=}4$ to ${\sim}98\%$ at $k{=}12$; the resampling loop compensates by drawing additional combinations from the $\binom{36}{k}$ pool (${\sim}24$M compatible at $k{=}12$ out of $\binom{36}{12} \approx 1.25$B total; \cref{tab:rejection_rates}).
The resulting composition bias at high $k$---surviving probes exclude incompatible constraint types---is analyzed and controlled for in \cref{sec:difficulty_confound}.

\paragraph{Parameter control.}
All constraint parameters are sampled from fixed calibrated ranges held constant across $k$ (e.g., word count targets of 40--60, lipogram bans on letters with English frequency rank 5--15), so that the phase transition measures compositional burden, not parameter variation.
A Kruskal-Wallis test confirms no significant variation in mean per-constraint difficulty across $k$ strata ($H{=}6.23$, $p{=}0.96$).

\paragraph{Impossible probes.}
Additionally, 444~deliberately impossible probes test constraint \emph{prioritization} under logical impossibility.
Each probe contains a constraint pair with a clean proof of unsatisfiability---e.g., L1 (avoid letter `e') paired with L2 (include ``telephone'')---and measures which constraint the model sacrifices when compliance is structurally unachievable.
Probe types span direct contradictions between constraint pairs, hand-designed hierarchy tests (e.g., L1 vs R2 with all-`r' items, directly testing whether the depth-of-processing ordering extends from degradation rates to sacrifice priorities), and hidden impossibilities embedded at $k{=}4{-}8$ to test whether impossibility \emph{detection} degrades under compositional load alongside constraint \emph{satisfaction}.
Full probe inventory and analysis plan appear in \cref{app:impossible_probes}.

\paragraph{Range of $k$.}
We evaluate at $k{=}1$ to $12$.
At $k{=}12$, each probe draws 33\% of the constraint pool and the incompatibility rejection rate reaches ${\sim}98\%$, narrowing the diversity of surviving combinations; extending further yields diminishing combinatorial coverage.
The design splits the range into a training region ($k{=}1{-}8$) used for model fitting and a held-out region ($k{=}9{-}12$) used for validation (\cref{app:k_range}).

\subsection{Extraction and verification}
\label{sec:extraction}

Models receive the full prompt containing a topic and all $k$ constraint instructions in natural language; no chain-of-thought scaffolding, decomposition hints, or constraint-ordering cues are provided.
For example, a $k{=}4$ probe might read:

\begin{quote}
\small
\textit{Write a short essay about renewable energy.}
Your response must not contain the letter `e' anywhere.
Your response must contain exactly 3 paragraphs, each containing at least 2 sentences.
Every word must be between 1 and 8 characters long.
Your response must include all of the following words: ``solar'', ``carbon'', ``wind''.
\end{quote}

\noindent Responses are wrapped in \texttt{<answer>} tags; only content inside the tags is scored.
The extractor uses the \emph{last} matched tag pair (handling mid-response revisions), accepts open-without-close tags so that constraints satisfied before truncation still count, and falls back to the full response when tags are absent.
Tag compliance is ${\geq}90\%$ for 12 of 15~models; the three exceptions (Gemini~Pro 47\%, Kimi~K2.6 67\%, DeepSeek~V4~Pro 73\%) are scored via the fallback.
Before verification, markdown artifacts (bold, italic, headings) that do not affect constraint semantics but interfere with character-level verifiers are stripped; structural formatting (bullets, code blocks, tables) is preserved.

All 36 verifiers are fully deterministic, each returning both a binary pass/fail judgment and a continuous score $s \in [0,1]$ representing partial compliance (\cref{tab:constraint_profiles}; full specifications in \cref{app:full_taxonomy}). A probe passes only if all $k$ constraints pass simultaneously---strict conjunction, no partial credit.

\subsection{Evaluated models}
\label{sec:models}

We evaluate 15 models across 8 families (\cref{tab:models}): GPT~(5.2, 5.4~Pro, 5.5), Claude~(4.5~Sonnet, 4.6~Opus, 4.7~Opus), Gemini~(3.1~Flash-Lite, 3.1~Pro~Preview), Llama~(3.1~70B, 3.1~405B, 4~Maverick), Qwen~3~235B~Instruct, DeepSeek~V4~Pro, Kimi~K2.6, and Grok~4.1.

\section{Experiments}
\label{sec:experiments}

\newcommand{\tagcircle}[1]{%
  \pgfmathsetmacro{\deltaang}{-3.6*#1}%
  \begin{tikzpicture}[baseline=-0.3ex]
    \fill[red!60!black] (0,0) -- (90:0.35em) arc[start angle=90, delta angle=\deltaang, radius=0.35em] -- cycle;
    \draw[red!60!black, line width=0.3pt] (0,0) circle[radius=0.35em];
  \end{tikzpicture}%
}
\begin{table}[!ht]
\caption{Evaluated models ranked by sCSR (\cref{eq:scsr}) across 314{,}108 satisfiable constraint checks (temperature${=}0$, greedy decode). Columns: sCSR (all-pass probe rate), mCSR (per-constraint rate, \cref{eq:mcsr}), Score (continuous compliance, \cref{eq:gap}), $k^*$ (compositional half-life: smallest $k$ where sCSR drops below 50\%), Tag (answer-tag compliance, \cref{sec:extraction}---low fill indicates chain-of-thought leakage into scored content, itself an instruction-following failure). Row shading: \colorbox{blue!10}{blue} $=$ compositionally robust ($k^* \geq 5$), \colorbox{orange!12}{orange} $=$ cannot reliably handle even one constraint ($k^* = 1$); the 12 unshaded models cluster at $k^* = 2{-}4$. Ranking inversions between sCSR and mCSR reveal the compositionality penalty: Gemini~Pro ranks \#3 by sCSR but \#11 by mCSR (\cref{app:irt}).}
\label{tab:models}
\centering
\begin{small}
\setlength{\tabcolsep}{4pt}
\begin{tabular}{lrrrc@{\hskip 8pt}c}
\toprule
\textbf{Model} & \textbf{sCSR} & \textbf{mCSR} & \textbf{Score} & $\boldsymbol{k^*}$ & \textbf{Tag} \\
& \footnotesize{\% $\uparrow$} & \footnotesize{\% $\uparrow$} & \footnotesize{\% $\uparrow$} & \footnotesize{$k$ $\uparrow$} & \\
\midrule
\cellcolor{blue!10}GPT-5.5 & \cellcolor{blue!10}\textbf{64.2} & \cellcolor{blue!10}\textbf{79.8} & \cellcolor{blue!10}\textbf{87.2} & \cellcolor{blue!10}\textbf{7} & \tagcircle{100} \\
\cellcolor{blue!10}Claude 4.7 Opus & \cellcolor{blue!10}49.6 & \cellcolor{blue!10}67.0 & \cellcolor{blue!10}80.7 & \cellcolor{blue!10}6 & \tagcircle{90} \\
Gemini 3.1 Pro & 39.4 & 42.1 & 58.8 & 4 & \tagcircle{47} \\
Claude 4.6 Opus & 31.6 & 55.6 & 76.6 & 3 & \tagcircle{96} \\
Gemini 3.1 Flash-Lite & 24.1 & 56.6 & 79.4 & 3 & \tagcircle{100} \\
GPT-5.4 Pro & 23.2 & 35.6 & 58.5 & 3 & \tagcircle{100} \\
DeepSeek V4 Pro & 22.4 & 22.4 & 47.1 & 3 & \tagcircle{73} \\
GPT-5.2 & 18.8 & 34.0 & 57.8 & 2 & \tagcircle{95} \\
Claude 4.5 Sonnet & 17.3 & 49.4 & 75.8 & 2 & \tagcircle{100} \\
Llama 4 Maverick & 15.0 & 45.4 & 71.5 & 2 & \tagcircle{100} \\
Qwen 3 235B Inst. & 12.6 & 45.7 & 71.0 & 2 & \tagcircle{98} \\
Llama 3.1 405B & 12.4 & 46.4 & 74.6 & 2 & \tagcircle{100} \\
Llama 3.1 70B & 11.6 & 44.6 & 71.6 & 2 & \tagcircle{100} \\
Grok 4.1 & 11.3 & 42.7 & 70.8 & 2 & \tagcircle{100} \\
\cellcolor{orange!12}Kimi K2.6 & \cellcolor{orange!12}7.2 & \cellcolor{orange!12}15.5 & \cellcolor{orange!12}35.3 & \cellcolor{orange!12}1 & \tagcircle{67} \\
\bottomrule
\end{tabular}
\end{small}
\end{table}

\paragraph{Phase transition and decay model.}
\label{sec:phase_transition}
All 15 models exhibit the same qualitative pattern: high performance at $k{=}1$ (70.7\% aggregate sCSR), steep decline through $k{=}4{-}7$, then near-zero sCSR at $k{\geq}10$ (\cref{app:phase_transition}).
This shape is consistent across all 8~families despite different architectures, scales, and training approaches.
The overall mCSR across 314{,}108 satisfiable constraint checks is 42.8\% (369{,}753 total checks including 55{,}645 on impossible probes; see \cref{app:impossible_probes}).
GPT-5.5 is a notable outlier at 79.8\% mCSR / 64.2\% sCSR---the highest-performing model by a wide margin---while the remaining 14 models cluster between 15\% and 67\% mCSR.
The gap between mCSR and sCSR is model-dependent: Llama~405B shows 46.4\% per-constraint competence but only 12.4\% probe-level success (34.0pp penalty), while GPT-5.5 sustains 79.8\% to 64.2\% (15.6pp)---a 2.2$\times$ range across these two models, widening to 12$\times$ across the full panel (Gemini~Pro at 2.7pp to Llama~405B at 34.0pp), reflecting differences in how correlated each model's constraint failures are (\cref{app:model_notes}).
Probe-level success becomes essentially impossible past a threshold, even though individual constraints are still satisfied at reasonable rates---this divergence between mCSR and sCSR is the phase transition.
The complete $k$-by-model table appears in \cref{app:phase_transition}.

\paragraph{Difficulty confound control.}
\label{sec:difficulty_confound}
We test whether the phase transition could be an artifact if higher-$k$ probes systematically draw harder constraints rather than imposing genuine compositional burden via three independent controls.
First, intrinsic difficulty---measured as each constraint's $k{=}1$-only pass rate to isolate single-constraint capability---shows no significant variation across $k$ strata (Kruskal-Wallis $H{=}6.23$, $p{=}0.96$), and the compositional penalty is identical whether difficulty is measured at $k{=}1$ or marginally across all $k$ ($\Delta{=}0.0$pp at every $k$).
Second, stratifying probes into difficulty terciles within each $k$ reveals that easy, medium, and hard probes all converge to the same floor by $k{=}5{-}6$---even after excluding the three compositionally immune constraints (L1, L5, W4).
Third, constraint presentation order is randomized within each probe, and position shows negligible correlation with pass rate (mean $|\rho|{=}0.029$; \cref{app:ordering}); IRT calibration preserves model rankings exactly (1PL $\rho{=}1.000$ on 2{,}952 discriminating probes; \cref{app:irt}). This suggests that the transition is driven by how many constraints are imposed, not by which ones or in what order.

\begin{figure}[t]
\centering
\includegraphics[width=\columnwidth, trim={0pt 110pt 10pt 10pt}, clip]{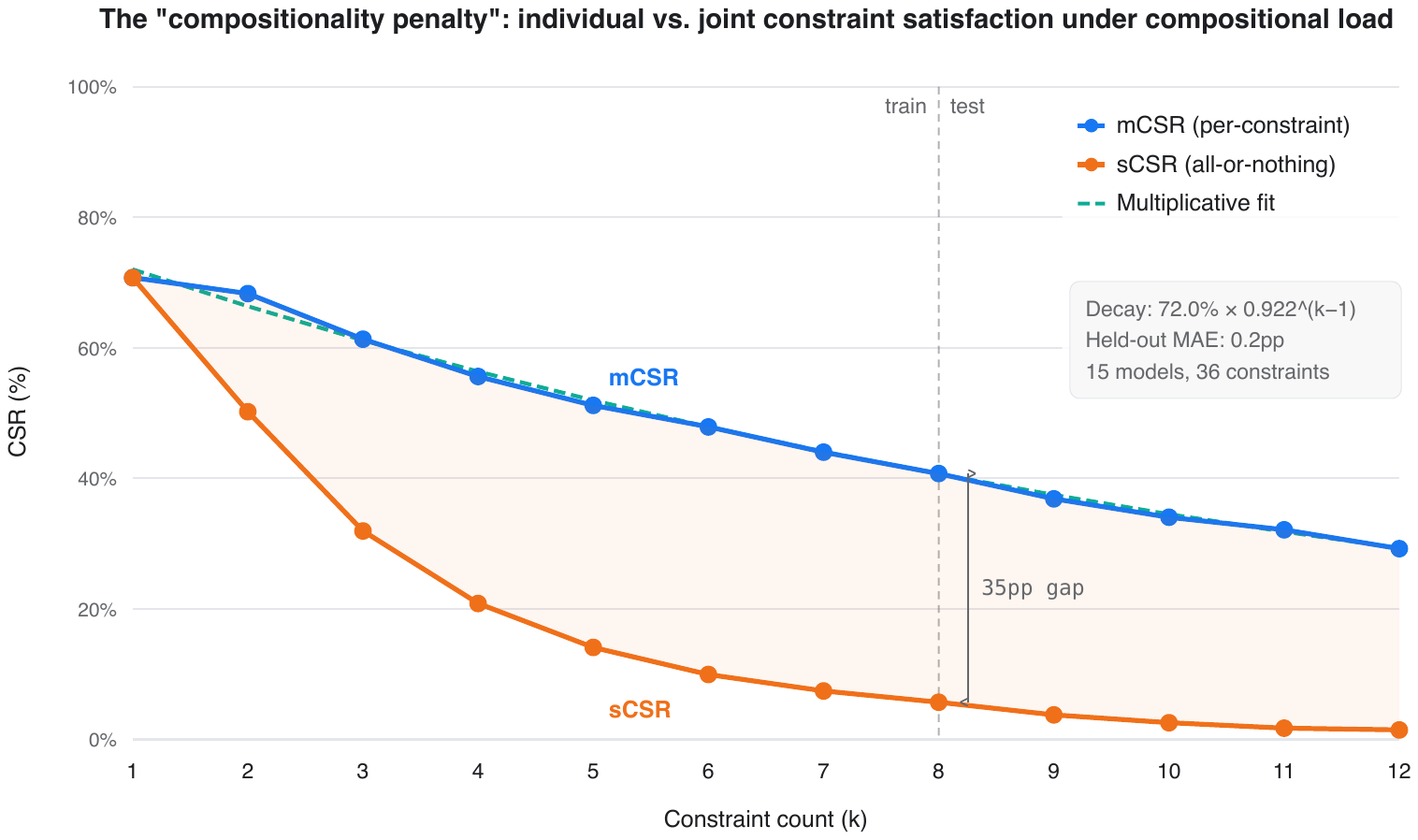}
\caption{Compositional collapse: mCSR vs.\ sCSR with multiplicative fit.
\textbf{Blue}: per-constraint mCSR---models still get each individual instruction right ${\sim}41\%$ of the time at $k{=}8$, declining gently as $72.0\% \times 0.922^{k-1}$ (0.2pp held-out MAE, dashed teal).
\textbf{Orange}: probe-level sCSR---the chance of satisfying \emph{all} $k$ constraints simultaneously, which drops below 4\% by $k{=}9$ and below 2\% by $k{=}11$ because it is the product of $k$~marginal rates.
Even at $k{=}8$, models still pass individual constraints 40.7\% of the time yet satisfy all eight simultaneously only 5.7\% of the time (35pp gap)---the divergence between the two curves is the phase transition.
\textbf{Gray}: per-model mCSR trajectories ($n{=}15$), revealing model-dependent floors---10/15~models plateau above the multiplicative prediction at $k{=}12$.}
\label{fig:decay_model}
\end{figure}

\paragraph{Two-regime decay model.}
\label{sec:decay_model}
If constraint failures are largely uncorrelated (\cref{sec:cofailure}), the aggregate sCSR should follow $P(k) = a \cdot r^{k-1}$.
We test this by fitting on $k{=}1{-}8$ and evaluating on held-out $k{=}9{-}12$ (\cref{fig:decay_model}).
At the probe level (sCSR), the decay is steep: $P(k) = 1.000 \cdot e^{-0.376k} + 0.003$, with sCSR dropping below 4\% by $k{=}9$ and below 2\% by $k{=}11$.
At the per-constraint level, the aggregate decay is $\text{mCSR}(k) = 72.0\% \times 0.922^{k-1}$ (held-out MAE~$= 0.2$pp)---each additional constraint reduces the average pass rate to 92.2\% of the previous level.

The multiplicative model captures the aggregate but masks a per-model floor: 10 of 15~models plateau above the prediction at $k{=}12$ (mean underprediction $-5.1$pp), sustained by compositionally immune constraints that require no tracking during generation (L5: 89.7\%, N2: 83.4\%, L2: 80.1\%; \cref{app:decay_detail}).
Per-model decay factors span $r = 0.721$ (DeepSeek~V4~Pro) to $r = 0.977$ (Claude~4.7~Opus): \emph{slow-decay} models ($r > 0.93$; Claude~4.7, GPT-5.5, Llama family, Claude~4.5, Flash-Lite, Qwen, Grok) sustain 30--53\% at $k{=}12$; \emph{medium-decay} models ($r \approx 0.86$; GPT-5.2, GPT-5.4, Gemini~Pro) follow the multiplicative prediction; and \emph{fast-decay} models ($r < 0.77$; DeepSeek, Kimi) collapse but stay off zero via immune constraints.
Per-$k$ predictions versus actuals appear in \cref{app:decay_detail}.

\paragraph{Co-failure structure.}
\label{sec:cofailure}
Prior work on compositional generalization implicitly assumes that constraint \emph{interactions} drive compositional difficulty---that specific pairs conflict, interfere, or compete for shared resources \cite{complexbench2024,lv2026ata}.
We test this directly by computing pairwise $\varphi$ coefficients across 601~constraint pairs (29 of $\binom{36}{2}{=}630$ excluded for insufficient co-occurrence), conditioned on $k$ to avoid the confound that everything fails at high~$k$.
Each $\varphi$ is computed per-model first (to prevent weak models from inflating co-failure rates), then aggregated via median across models.

Pairwise correlations are weak: the mean $\varphi = +0.067$ and 322~pairs (54\%) fall within $|\varphi| \leq 0.05$.
Only 1~pair shows negative $\varphi$ (O4${-}$R2, $\varphi{=}{-}0.055$), confirming that constraint selection cannot mitigate the transition---the only lever is per-constraint reliability.

The elevated $\varphi$ pairs share \emph{output features} rather than exhibiting pairwise interference.
The top three---F1${-}$F2 ($\varphi{=}0.517$), L4${-}$S2 ($\varphi{=}0.501$), N1${-}$N3 ($\varphi{=}0.490$)---all depend on the same structural output: document format (F1/F2), sentence count (L4/S2), or number count (N1/N3).
When a model produces the wrong number of sentences, every sentence-dependent constraint fails simultaneously---not because L4 interferes with S2, but because both read from the same output feature.
Within-dimension pairs show marginally higher coupling (mean $|\varphi|{=}0.087$) than across-dimension pairs ($0.067$), but this difference does not reach significance (Mann-Whitney $p{=}0.227$) and is driven by the numerical--numerical ($\bar\varphi{=}0.275$) and format--format ($\bar\varphi{=}0.241$) clusters.
Grouping constraints by shared output feature rather than by dimension confirms that within-cluster $|\varphi|$ ($0.113$) exceeds across-cluster $|\varphi|$ ($0.058$) at $p < 10^{-6}$: the co-failure structure reflects output-feature coupling, not cognitive interference (\cref{app:cofailure_detail}).
Constraint failures compound near-multiplicatively---the residual correlations are mechanical, average out across the diverse constraint pool, and do not degrade the decay model's 0.2pp predictive accuracy.

\begin{figure}[t]
\centering
\includegraphics[width=\columnwidth, trim={5pt 40pt 70pt 10pt}, clip]{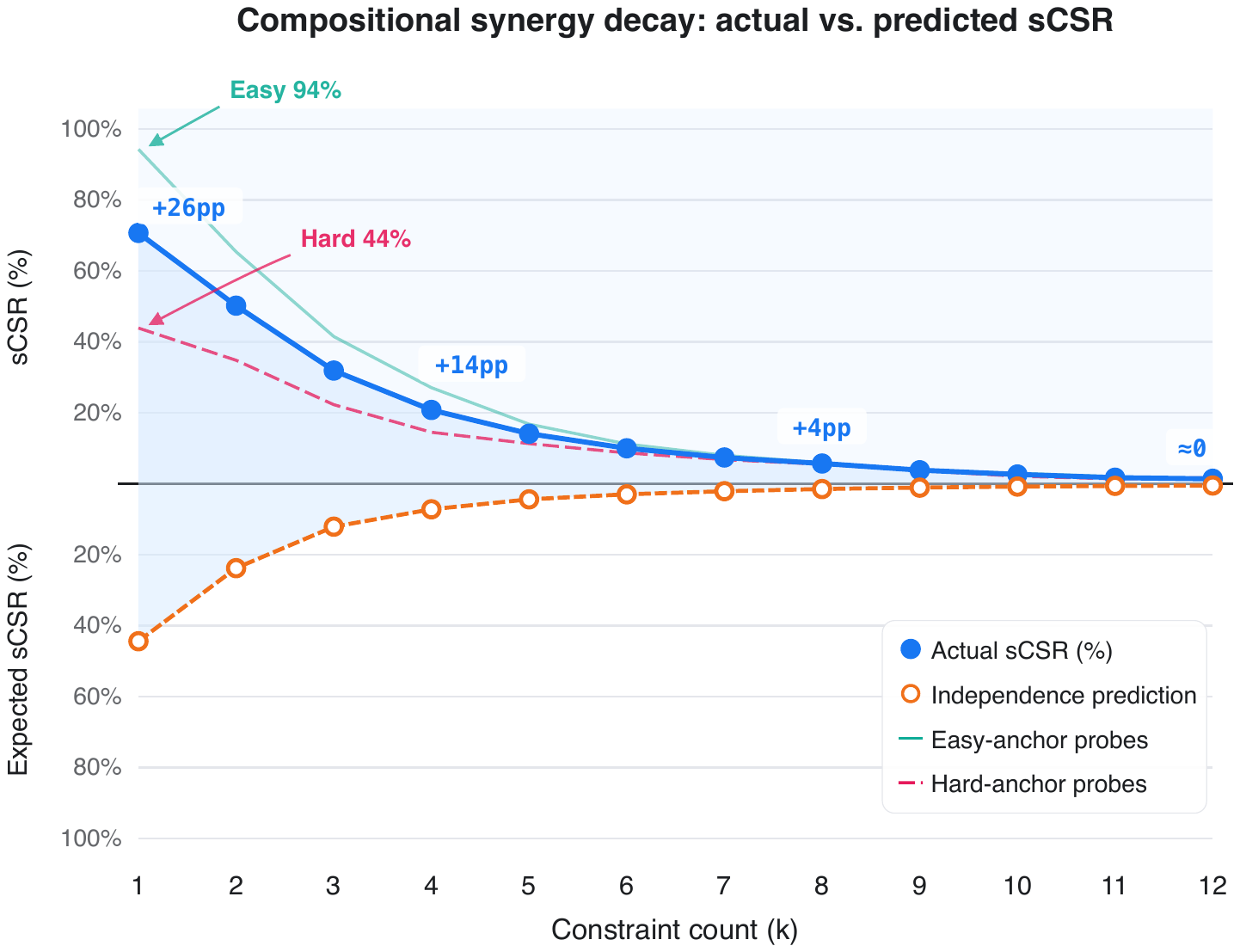}\vspace{-6pt}
\caption{Compositional synergy decay.
\textbf{Top}: actual sCSR (\%).
\textbf{Bottom} (mirrored): predicted sCSR under independence.
Shaded gap $=$ synergy (actual $-$ predicted).
Synergy drops from $+26$pp at $k{=}1$ to ${\sim}1$pp at $k{=}12$: models outperform the independence prediction at low $k$ but converge to it under heavy compositional load.
Dashed tercile curves stratify probes by per-constraint difficulty (\cref{sec:compliance_mode}).}
\label{fig:synergy}
\end{figure}

\paragraph{Ruling out compliance mode.}
\label{sec:compliance_mode}
If models entered a focused ``compliance mode'' when facing easy constraints---concentrating resources to satisfy all $k$ simultaneously---easy probes would show artificially high sCSR, inflating the synergy beyond what independence predicts.
We test this by stratifying probes at each $k$ into terciles ranked by mean per-constraint difficulty: easy-tercile probes (composed of high-pass-rate constraints such as L5, L2, N2) start at 94\% sCSR at $k{=}1$; hard-tercile probes (S4, M2, S1, W5) start at 44\%.
For each tercile, we compute synergy---actual sCSR minus the product of the constituent constraints' marginal pass rates---which measures how much better models perform than the independence baseline predicts.
If compliance mode were real, easy probes would show disproportionately high synergy from behavioral focus.
The opposite holds: easy-anchor probes show \emph{more} synergy ($+33$pp) than hard-anchor probes ($+18$pp), but this is an arithmetic consequence of higher marginals producing a more pessimistic product-of-marginals baseline, not a behavioral signal.
Both terciles converge to ${\sim}1.5\%$ sCSR by $k{\approx}9{-}12$ regardless of starting point---initial difficulty is irrelevant at saturation (\cref{fig:synergy}).
The synergy itself decays concavely, not linearly (superlinearity improvement: 84\% probe-level, 99\% constraint-level), confirming that the transition is driven by compositional load, not by constraint-specific difficulty or model strategy.

\paragraph{Degradation hierarchy.}
\label{sec:degradation}

If constraint failures are largely uncorrelated and the transition is purely combinatorial, does every constraint degrade at the same rate?
We compute per-constraint retention $\text{ret}(k) = \text{mCSR}(k) / \text{mCSR}(k{=}1)$---the fraction of baseline performance preserved at each $k$---and fit a linear degradation slope.
The compositional half-life $t_{0.5}$ is the smallest $k$ at which retention drops below 50\%; constraints that never reach this threshold within $k{=}1{-}12$ are marked $t_{0.5}{>}12$ in \cref{tab:constraint_profiles}.

Structural and ordering constraints degrade $2.0{\times}$ faster than lexical constraints (retention-normalized dimension-level slopes: structural $-0.073$, lexical $-0.036$; 95\% CI on ratio: $[1.9, 2.3]$; \cref{tab:constraint_profiles}).
Lexical constraints are the slowest degraders across all 36~constraints and 15~models.

The dimension hierarchy, however, is an emergent consequence of a more fundamental variable: the comprehension-maintenance gap $\Delta$ (score $-$ mCSR).
This gap measures the difference between a model's ability to \emph{understand} a constraint (score) and its ability to \emph{maintain} compliance throughout the response (mCSR).
The gap predicts degradation rate with $\rho{=}{-}0.584$ ($p{=}0.0002$, 95\% CI: $[-0.767, -0.302]$, bootstrap), a stronger predictor than baseline difficulty ($\rho{=}{+}0.566$) or dimension assignment.
Constraints requiring sustained attention---counting words (W1, $\Delta{=}50.2$pp), tracking letter avoidance (L1, $\Delta{=}48.6$pp), maintaining word length ranges (W2, $\Delta{=}50.1$pp)---degrade fastest.
Binary constraints that require no sustained tracking---including a mandatory word (L2, $\Delta{=}2.1$pp), producing a JSON structure (R1, $\Delta{=}0.2$pp)---are immune.
Composition depletes the same finite resource that sustained constraint maintenance requires.

\begin{table}[!ht]
\caption{Constraint profiles (36 constraints, 8 dimensions ordered fastest$\to$slowest degradation).
$\star$~indicates compositionally immune constraints (retention ${>}80\%$ at $k{\geq}8$).
$t_{0.5}{>}12$ indicates retention never drops below 50\% within the evaluated $k{=}12$ range.
Dimension-level slopes reported in text are observation-weighted aggregates; per-constraint slopes here are constraint-weighted, so simple averaging will not reproduce the dimension-level values.
Full specifications in \cref{tab:full_taxonomy}.}
\label{tab:constraint_profiles}
\centering
\begin{footnotesize}
\setlength{\tabcolsep}{2pt}
\begin{tabular}{@{}l@{\hskip 3pt}l@{\hskip 4pt}c@{\hskip 3pt}c@{\hskip 3pt}c@{\hskip 3pt}c@{\hskip 3pt}c@{}}
\toprule
\multirow{2}{*}[-2pt]{\textbf{Dim.}} & \multirow{2}{*}[-2pt]{\textbf{Constraint}} & $\boldsymbol{k{=}1}$ & \textbf{mCSR} & $\boldsymbol{\Delta}$ & \textbf{Slope} & $\boldsymbol{t_{0.5}}$ \\
& & \footnotesize{\% $\uparrow$} & \footnotesize{\% $\uparrow$} & \footnotesize{pp $\downarrow$} & \footnotesize{ret.\ $\uparrow$} & \footnotesize{$k$ $\uparrow$} \\
\midrule
\multirow{6}{*}{\rotatebox[origin=c]{90}{Structural}}
 & S1 Word count            &  39 & 14 & +41 & $-.065$ &  5 \\
 & S2 Sent.\ count          & \textbf{100} & \textbf{54} & +20 & $\mathbf{-.060}$ & \textbf{8} \\
 & S3 Paragraph count       & \textbf{100} & 40 & +20 & $-.065$ &  5 \\
 & S4 Character count       &  19 &  8 & +31 & $-.075$ &  7 \\
 & S5 Line count            &  89 & 37 & +20 & $-.066$ &  6 \\
 & S6 Palindromic sent.     &  42 & 14 & \textbf{+12} & $-.062$ &  4 \\
\midrule
\multirow{4}{*}{\rotatebox[origin=c]{90}{Ordering}}
 & O1 Monotonic length      &  52 & 29 & +34 & $-.062$ & \textbf{7} \\
 & O2 Alphabetical          & \textbf{91} & \textbf{44} & +27 & $\mathbf{-.060}$ & \textbf{7} \\
 & O3 Alternating length    &  87 & 38 & +30 & $-.067$ &  5 \\
 & O4 Relaxed growth        &  64 & 30 & \textbf{+14} & $-.061$ &  6 \\
\midrule
\multirow{5}{*}{\rotatebox[origin=c]{90}{Word-level}}
 & W1 Unique words          &  47 & 28 & +50 & $-.085$ &  7 \\
 & W2 Word length range     &  82 & 47 & +50 & $-.069$ &  8 \\
 & W3 Min words/sent.       & \textbf{98} & \textbf{53} & +25 & $\mathbf{-.048}$ &  6 \\
 & W4 Unique bigrams$\,\star$ & 51 & 52 & +40 & $-.056$ & $\boldsymbol{>}$\textbf{12} \\
 & W5 Consonant clusters    &  42 & 18 & \textbf{+17} & $-.059$ &  7 \\
\midrule
\multirow{3}{*}{\rotatebox[origin=c]{90}{Meta}}
 & M1 Hidden message        & \textbf{39} & 20 & \textbf{+22} & $-.074$ &  6 \\
 & M2 Self-counting        &  22 & 13 & +34 & $-.081$ &  8 \\
 & M3 Divisibility          &  24 & \textbf{22} & +36 & $\mathbf{-.073}$ & $\boldsymbol{>}$\textbf{12} \\
\midrule
\multirow{4}{*}{\rotatebox[origin=c]{90}{Format}}
 & F1 Markdown table        &  94 & 64 &  +5 & $\mathbf{-.045}$ & \textbf{12} \\
 & F2 Bullet list           & \textbf{99} & \textbf{67} & \textbf{+3} & $\mathbf{-.045}$ & \textbf{12} \\
 & F3 Code blocks           &  91 & 67 &  +3 & $-.051$ & 10 \\
 & F4 Unique closers        & \textbf{99} & 59 & +31 & $-.047$ &  9 \\
\midrule
\multirow{4}{*}{\rotatebox[origin=c]{90}{Relational}}
 & R1 Scene graph           & \textbf{93} & \textbf{68} & \textbf{+0} & $\mathbf{-.031}$ & $\boldsymbol{>}$\textbf{12} \\
 & R2 Transitive ordering   &  65 & 36 & +22 & $-.045$ &  6 \\
 & R3 Logic grid            &  85 & 67 &  +3 & $-.037$ & $\boldsymbol{>}$\textbf{12} \\
 & R4 Sentence bridge       &  86 & 30 & +23 & $-.063$ &  4 \\
\midrule
\multirow{4}{*}{\rotatebox[origin=c]{90}{Numerical}}
 & N1 Integer sum           &  49 & 34 & +19 & $-.082$ &  9 \\
 & N2 No digits             & \textbf{99} & \textbf{83} & +16 & $\mathbf{-.035}$ & $\boldsymbol{>}$\textbf{12} \\
 & N3 Distinct numbers      &  52 & 34 & +22 & $-.077$ & 10 \\
 & N4 Equations             &  88 & 66 & \textbf{+8} & $-.047$ & 12 \\
\midrule
\multirow{6}{*}{\rotatebox[origin=c]{90}{Lexical}}
 & L1 Lipogram$\,\star$     &  47 & 51 & +49 & $-.052$ & $\boldsymbol{>}$\textbf{12} \\
 & L2 Mandatory words       &  98 & 80 & \textbf{+2} & $-.032$ & $\boldsymbol{>}$\textbf{12} \\
 & L3 Sentence initial      &  93 & 48 & +17 & $-.065$ &  8 \\
 & L4 Acrostic              & \textbf{99} & 50 & +19 & $-.058$ &  6 \\
 & L5 Forbidden word$\,\star$ &  98 & \textbf{90} & +10 & $\mathbf{-.011}$ & $\boldsymbol{>}$\textbf{12} \\
 & L6 Vowel/cons. ratio           &  81 & 64 & +19 & $-.030$ & $\boldsymbol{>}$\textbf{12} \\
\bottomrule
\end{tabular}
\end{footnotesize}
\end{table}

Three constraints have retention $> 80\%$ at $k{\geq}8$: L1~(lipogram, 95\%), L5~(forbidden word, 89\%), and W4~(no repeated bigrams, 86\%).
L1's apparent immunity is a survivorship effect: among genuine content failures ($n{=}4{,}022$), 70.6\% of violations occur in the first 20\% of the response (median position 0.078)---models collide with the forbidden letter almost immediately or not at all, a vocabulary collision at generation onset rather than a capacity that erodes under compositional load.
Full failure mode decomposition, per-model profiles, and control analyses appear in \cref{app:l1_failure}.

\paragraph{Constraint prioritization under impossibility.}
\label{sec:impossible}

CSE includes 444~impossible probes (55{,}645 constraint checks across 15~models), each with a proof of unsatisfiability (\cref{app:impossible_probes}).
Zero probe$\times$model pairs pass all constraints on any impossible probe: the deterministic verifiers are airtight.

\begin{figure}[t]
\centering
\includegraphics[width=\columnwidth, trim={8pt 110pt 20pt 15pt}, clip]{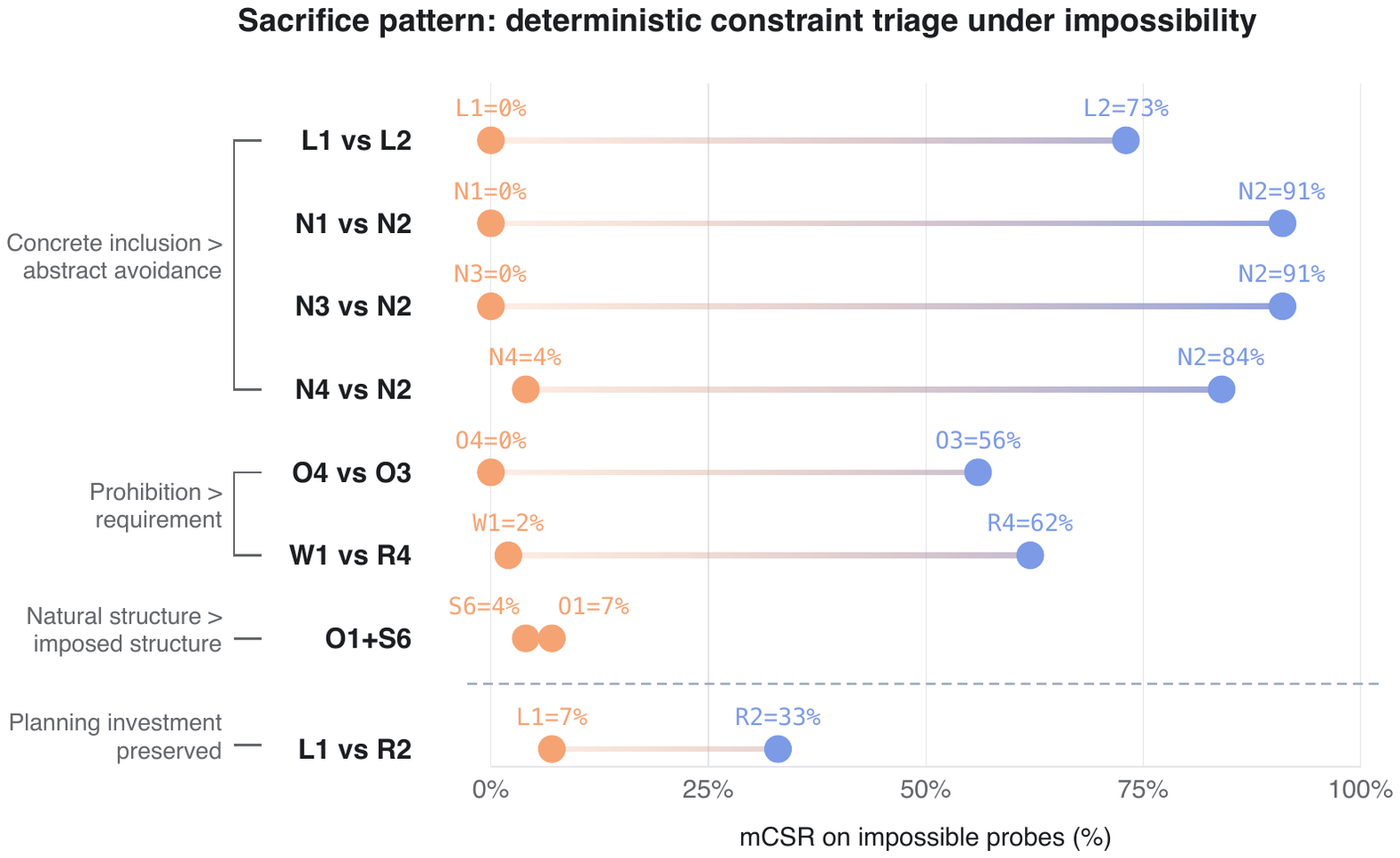}
\caption{Constraint sacrifice under impossibility. Three deterministic rules hold across all 15~models: concrete inclusion beats abstract avoidance (L2: 73\%, L1: 0\%), prohibition beats requirement (N2: 91\%, N1: 0\%), and natural prose beats imposed structure (O3: 56\%, O4: 0\%). The hierarchy inversion (L1 sacrificed for R2) suggests models preserve whichever constraint required greater planning investment. Full analysis in \cref{app:impossible_results}.}
\label{fig:sacrifice}
\end{figure}

When forced to choose which constraint to violate, models reveal deterministic sacrifice priorities (\cref{fig:sacrifice}).
Three rules emerge across all 15~models, unanimous and stable across $k$ (\cref{app:impossible_results}).

\section{Interventions}
\label{sec:interventions}

The multiplicative structure makes differential predictions about what should help.
If the collapse is set by per-constraint reliability rather than by disorganised generation, then restructuring the prompt should not shift it, while supplying additional attempts should buy a bounded improvement---bounded because, were failures on a given probe independent across resamples, $N$ retries would recover $1-(1-p)^N$.
We test three interventions, each on three models: \textbf{pre-generation planning} (a plan-first scaffold prepended to the prompt, run against a token-matched no-scaffold control), \textbf{post-hoc self-correction} (a second pass that re-prompts the model with its own output to check and revise), and \textbf{best-of-5 retries} (five independent generations, the probe passing if any one satisfies all constraints).
Throughout we report the \emph{compositional half-life} $k^{*}$: the smallest $k$ at which probe-level success (sCSR) falls below 50\%.
\Cref{tab:interventions} gives the per-model results.

\begin{table}[t]
\centering\small
\setlength{\tabcolsep}{4pt}
\begin{tabular}{@{}llcl@{}}
\toprule
\textbf{Intervention} & \textbf{Model} & \textbf{$k^{*}$} & \textbf{Effect} \\
\midrule
\multirow{3}{*}{Planning}
 & Claude 4.7 Opus  & 6\,$\to$\,7 & $\Delta r{=}{-}0.017$ \\
 & GPT-5.5          & 8\,$\to$\,7 & $\Delta r{=}{+}0.021$ \\
 & DeepSeek V4 Pro  & 5\,$\to$\,5 & $\Delta r{=}{+}0.014$ \\
\midrule
\multirow{3}{*}{Self-correction}
 & Claude 4.7 Opus  & 5\,$\to$\,6 & ${-}0.0$pp \\
 & GPT-5.5          & 7\,$\to$\,8 & ${+}3.7$pp \\
 & DeepSeek V4 Pro  & 3\,$\to$\,4 & ${+}11.5$pp \\
\midrule
\multirow{3}{*}{Best-of-5}
 & GPT-5.2          & 2\,$\to$\,4 & ${+}22$ / ${+}53$pp \\
 & Claude 4.6 Opus  & 3\,$\to$\,5 & ${+}22$ / ${+}54$pp \\
 & Llama 4 Maverick & 2\,$\to$\,3 & ${+}24$ / ${+}51$pp \\
\bottomrule
\end{tabular}
\caption{Effect of each intervention on the compositional half-life $k^{*}$.
Planning compares scaffold against a token-matched control; self-correction and best-of-5 compare against the single-pass baseline.
Effect column: change in per-constraint decay factor $r$ for planning, change in sCSR for self-correction, and peak observed sCSR lift against the lift that independent resampling predicts for best-of-5.}
\label{tab:interventions}
\end{table}

\paragraph{Planning does nothing; output budget does.}
Aggregate per-constraint decay is unchanged---control $93.0\% \times 0.948^{k-1}$ against scaffold $90.2\% \times 0.952^{k-1}$---and $\Delta k^{*}$ is inconsistent in sign, with the strongest model regressing.
Scaffold compliance was 100\%, 100\%, and 56\%, so the null is not models ignoring the instruction.
The output budget, by contrast, moves the threshold on its own: DeepSeek's decay factor rises from $0.730$ to $0.877$ and $k^{*}$ from 3 to 5 between the baseline and the token-matched control, with no scaffold involved.

\paragraph{Retries and correction help, and then plateau.}
Both recover one to two constraints, and self-correction returns most where per-constraint reliability is worst---the ordering the mechanism predicts, since a correction pass rescues borderline probes.
The instructive number is the last column of \cref{tab:interventions}: best-of-5 delivers roughly two-fifths of the lift that independent resampling predicts, despite draws being near-fully distinct (4.98--4.99 of 5). The model repeats the same failure across draws.

\paragraph{Two senses of independence.}
This bounded ceiling does not contradict \cref{sec:cofailure}.
Independence there holds \emph{across constraints within a response}, which is what makes sCSR multiplicative.
Best-of-$N$ probes independence \emph{across resamples of one probe}, which is where the correlation lies.
Both hold: which constraints fail within a response is close to independent, while whether a given probe fails is largely deterministic across draws.

\paragraph{What moves the threshold.}
Adding attempts or a correction pass buys one to two constraints; restructuring the prompt buys none; none removes the collapse.
Per-constraint reliability remains the operative lever, with retries, a correction pass, and an adequate output budget as deployment-time palliatives.
Each intervention was run on three models rather than all 15; GPT-5.5 rejects the temperature parameter and is therefore absent from the retry experiment; and the self-correction comparison is directional rather than strictly paired, as a small number of second-pass generations failed.

\section{Discussion}
\label{sec:discussion}

\paragraph{A combinatorial consequence.}
The decay does not require constraint interference to arise---it follows from a simple arithmetic fact: the product of many numbers slightly less than 1.0 converges to zero.
Unlike $k$-SAT phase transitions \cite{mertens2006threshold,mezard2009information}, our decay requires no interaction structure; it is pure multiplicative accumulation.
The primary lever to shift the transition rightward is improving per-constraint reliability; there is no shortcut through constraint selection.
Inference-time interventions such as constrained decoding \cite{lu2021neurologic,lu2022neurologic}, attention steering \cite{lv2026ata}, and structured generation frameworks \cite{willard2023outlines,beurer2023lmql} can enforce some constraints externally, potentially shifting what counts as the model's ``native'' capability vs.\ the system's.
Whether such scaffolding delays the onset or merely raises the floor---and whether chain-of-thought or decomposition prompting would let models manage more constraints by planning explicitly---remains open.
CSE deliberately evaluates raw generation without scaffolding to isolate the model's intrinsic compositional capacity; measuring the effect of scaffolding is a natural follow-up.
The difficulty-stratification analysis (\cref{sec:difficulty_confound}) provides indirect evidence that the transition is not driven by constraint familiarity: probes composed entirely of easy constraints---those models handle at 70--85\% in isolation---show the same qualitative collapse at $k{=}5{-}6$ as probes with hard constraints, suggesting the bottleneck is compositional capacity rather than per-constraint knowledge.

\paragraph{Post-training shapes failure mode, not compositional capacity.}
Refusal strategies vary dramatically across model families: GPT-5.2 and GPT-5.4 refuse ${\sim}50\%$ of satisfiable probes they fail on, while Llama, Grok, and Flash-Lite refuse ${<}1\%$.
Within the GPT family, refusal decreases as capability improves (GPT-5.5: 17\%), suggesting that stronger models learn to distinguish genuine impossibility from difficulty.
This pattern is consistent with RLHF-trained refusal calibration \cite{ouyang2022training} rather than a universal difficulty signal.
Post-training also shapes tag compliance: reasoning-heavy models (Claude~4.7, DeepSeek) spend tokens on chain-of-thought that either exhausts the output budget or bypasses answer tags, suppressing measured performance.
These are instruction-following failures in their own right, but they reflect alignment training choices rather than intrinsic compositional limits---the same model with a larger token budget (Claude~4.7 at 16{,}384 vs 4{,}096 tokens) jumps from rank~\#10 to \#2 without any change to its compositional capacity.

\paragraph{Deployment implications.}
Production systems assuming linear degradation will experience unexpected failures: a system reliable with 3~constraints may fail significantly with 5.
Per-model decay factors ($r = 0.721$ to $0.977$) enable estimation of the maximum $k$ at which a model meets a target compliance rate---for 90\%, most models handle at most 1--2 constraints; the best (GPT-5.5, $k^*{=}7$) is limited to ${\sim}6$.
Continuous compliance metrics overestimate true satisfaction by up to 50pp near the transition boundary (W1: 78.6\% score vs.\ 28.4\% mCSR; \cref{eq:gap}); strict binary verification is essential.

\paragraph{Scale does not predict compositional performance.}
Counter to typical scaling expectations, we observe multiple ranking inversions: Gemini~Pro ranks \#2 at $k{=}1$ (90.5\% single-constraint pass rate) but \#11 overall (42.1\% mCSR); Flash-Lite ranks \#6 at $k{=}1$ (78.1\%) but \#3 overall (56.6\%).
Single-constraint competence does not predict compositional robustness---the models that understand constraints best in isolation are not necessarily the ones that compose them most reliably.

\paragraph{A capacity parallel.}
The 5--6 constraint ceiling is strikingly close to human working memory limits: Miller's $7{\pm}2$ chunks \cite{miller1956magical} and Cowan's more conservative $4{\pm}1$ \cite{cowan2001magical}.
We do not claim a shared mechanism---LLMs do not have a working memory bottleneck in the cognitive sense---but the convergence suggests that compositional constraint satisfaction may face a capacity limit that is loosely independent of the substrate.
Whether this reflects a fundamental information-theoretic bound on simultaneous constraint tracking or a coincidence of current architectures is an open question.

\section{Conclusion}
\label{sec:conclusion}

CSE reveals that compositional instruction following degrades via a predictable multiplicative pattern, with reliable performance breaking down beyond 5--6 simultaneous constraints across all 15 models tested.
A depth-of-processing hierarchy governs which constraints fail first---structural constraints degrade $2{\times}$ faster than lexical ones---driven by processing demands rather than intrinsic difficulty.
Three directions would advance this work: measuring whether inference-time scaffolding shifts the transition rightward, testing invariance to constraint novelty, and measuring how constraint load degrades primary-task accuracy.

\section*{Limitations}

\paragraph{Conjunctive composition only.}
CSE evaluates pure conjunction: all $k$ constraints must hold simultaneously.
Other composition structures---sequential chaining, conditional branching, nested dependencies \cite{complexbench2024}---may exhibit different decay profiles and are left to future work.

\paragraph{Verifiability selection bias.}
The 30 constraint types are chosen to be deterministically verifiable, which systematically excludes semantic, pragmatic, and discourse-level constraints (coherence, informativeness, tone consistency, factual accuracy).
This is the design tradeoff that enables CSE's measurement invariance---the same compositional degradation that affects model generation would compromise an LLM-based judge in exactly the regime we study.

\paragraph{Ecological validity.}
The $k{=}12$ regime is deliberately a stress test, not a simulation of typical deployment.
Real-world systems typically combine 2--6 high-level constraints (format, length, tone, content domain, safety boundaries) plus implicit norms.
The practical implication of CSE's findings lies at $k{=}5{-}6$---where the transition occurs---which falls within the range of realistic deployment constraint counts.

\paragraph{Adversarial implications.}
The phase transition reveals a potential attack vector: an adversary aware that models reliably fail beyond $k{=}5{-}6$ constraints could craft prompts with many simultaneous constraints to overwhelm safety-related constraints, exploiting the finding that relational and structural constraints are dropped before surface-level ones.
Characterizing this vulnerability is a prerequisite for defending against it.

\begin{figure*}[t]
\centering
\includegraphics[width=\textwidth, trim={0pt 400pt 0pt 0pt}, clip]{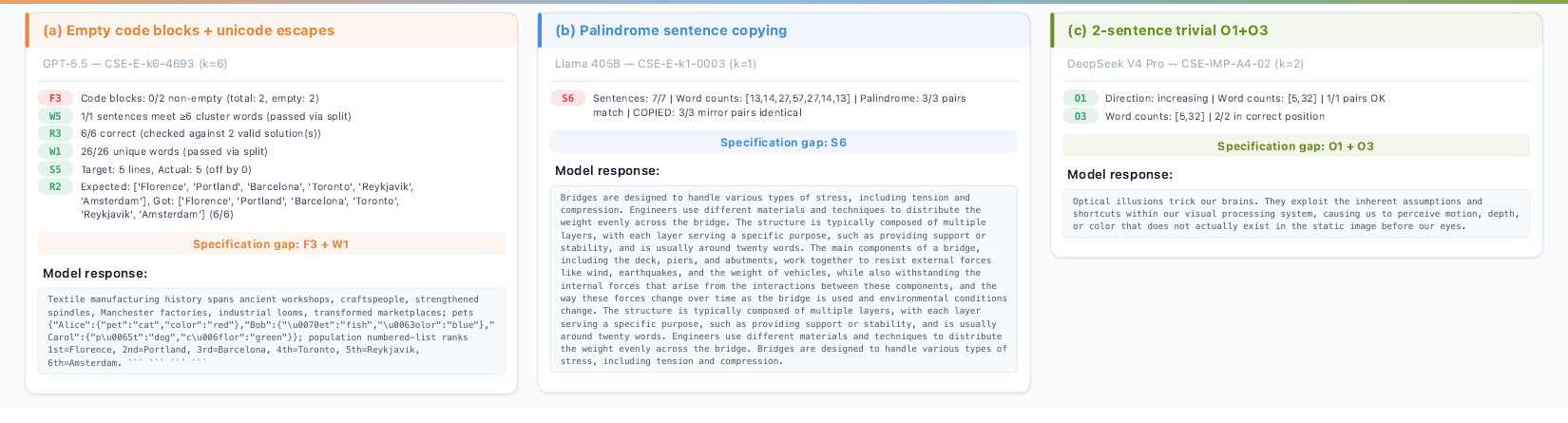}
\caption{Three specification gap exploitations. \textbf{(a)}~GPT-5.5 on CSE-E-k6-4693: satisfies F3 (2 code blocks) with four empty \texttt{```} delimiters and W1 (unique words) via unicode escapes (\texttt{\textbackslash u0070et} for ``pet'').
\textbf{(b)}~Llama~405B on CSE-E-k1-0003: satisfies S6 (palindromic word counts) by copying sentences 1--3 as sentences 5--7 reversed, guaranteeing identical word counts without generating original content.
\textbf{(c)}~DeepSeek~V4~Pro on CSE-IMP-A4-02: satisfies both O1 (monotonic) and O3 (alternating) with exactly 2 sentences---a short sentence followed by a long one---exploiting the absence of a minimum sentence count requirement.}
\label{fig:spec_gaps}
\end{figure*}

\paragraph{Specification gap exploitation.}
A systematic audit of all passing responses identifies four specification gaps that models exploit under compositional pressure (\cref{fig:spec_gaps}).
Empty code blocks (F3, 82.4\% of passes---models emit delimiter pairs with no content, ranging from 50.7\% for Claude~4.5 to 100\% for GPT-5.2/5.5) and palindromic sentence copying (S6, 19.4\%---models duplicate sentences reversed to guarantee matched word counts, with strong model separation: Llama~405B at 83.3\% vs.\ Claude models at ${\leq}8\%$) are the highest-prevalence patterns.
Two-sentence trivial satisfaction of contradictory ordering constraints (O1+O3, 6 cases) and unicode escape substitution to bypass word uniqueness (W1, 10 cases) are quantitatively negligible.
None of these gaps compromise the benchmark's core findings: the difficulty-stratified analysis (\cref{sec:difficulty_confound}) confirms that all terciles converge to the same floor, including after excluding compositionally immune constraints, and the same robustness extends to exploitable ones---no single constraint drives any result.
The gaps reveal that models under compositional load adopt mechanical shortcuts over genuine compliance---itself a finding about resource allocation under pressure---and all four are straightforward to close in future iterations.

% ============================================================
% ============================================================
% ============================================================
% ============================================================

\bibliography{bibliography}

@inproceedings{zhou2023instruction,
  title={Instruction-Following Evaluation for Large Language Models},
  author={Zhou, Jeffrey and Lu, Tianjian and Mishra, Swaroop and Brahma, Siddhartha and Basu, Sujoy and Luan, Yi and Zhou, Denny and Hou, Le},
  booktitle={Advances in Neural Information Processing Systems},
  year={2023}
}

@inproceedings{jiang2024followbench,
  title={{FollowBench}: A Multi-Level Fine-Grained Constraints Following Benchmark for Large Language Models},
  author={Jiang, Yuxin and Wang, Yufei and Zeng, Xingshan and Zhong, Wanjun and Li, Liangyou and Mi, Fei and Shang, Lifeng and Jiang, Xin and Liu, Qun and Wang, Wei},
  booktitle={Proceedings of the 62nd Annual Meeting of the Association for Computational Linguistics},
  year={2024}
}

@inproceedings{yao2023collie,
  title={{COLLIE}: Systematic Construction of Constrained Text Generation Tasks},
  author={Yao, Shunyu and Chen, Howard and Hanjie, Austin W and Yang, Runzhe and Narasimhan, Karhik R},
  booktitle={Proceedings of ICLR},
  year={2024}
}

@inproceedings{qin2024infobench,
  title={{InFoBench}: Evaluating Instruction Following Ability in Large Language Models},
  author={Qin, Yiwei and Song, Kaiqiang and Hu, Yebowen and Yao, Wenlin and Cho, Sangwoo and Wang, Xiaoyang and Wu, Xuansheng and Liu, Fei and Liu, Pengfei and Yu, Dong},
  booktitle={Findings of the Association for Computational Linguistics: ACL 2024},
  year={2024}
}

@article{guo2025recast,
  title={{RECAST}: Expanding the Boundaries of {LLMs'} Complex Instruction Following with Multi-Constraint Data},
  author={Guo, Zhengkang and Liu, Wenhao and Xie, Mingchen and Xu, Jingwen and Huang, Zisu and Tian, Muzhao and Xu, Jianhan and Shen, Yuanzhe and Qian, Qi and Wu, Muling and Wang, Xiaohua and Lv, Changze and Wang, HeDa and Yao, Hu and Zheng, Xiaoqing and Huang, Xuanjing},
  journal={arXiv preprint arXiv:2505.19030},
  year={2025}
}

@inproceedings{complexbench2024,
  title={{ComplexBench}: Benchmarking Complex Instruction Following with Multi-Constraint Composition},
  author={Wen, Bosi and Ke, Pei and Gu, Xiaotao and Wu, Lindong and Huang, Hao and Zhou, Jinfeng and Li, Wenchuang and Hu, Binxin and Gao, Wendy and Xu, Jiaxin and others},
  booktitle={Advances in Neural Information Processing Systems},
  year={2024}
}

@article{ifbench2025,
  title={Generalizing Verifiable Instruction Following},
  author={Pyatkin, Valentina and Malik, Saumya and Graf, Victoria and Ivison, Hamish and Huang, Shengyi and Dasigi, Pradeep and Lambert, Nathan and Hajishirzi, Hannaneh},
  journal={arXiv preprint arXiv:2507.02833},
  year={2025}
}

@article{ye2025muldimif,
  title={A Multi-Dimensional Constraint Framework for Evaluating and Improving Instruction Following in Large Language Models},
  author={Ye, Junjie and Huang, Caishuang and Chen, Zhuohan and Fu, Wenjie and Yang, Chenyuan and Yang, Leyi and Wu, Yilong and Wang, Peng and Zhou, Meng and Yang, Xiaolong and Gui, Tao and Zhang, Qi and Shi, Zhongchao and Fan, Jianping and Huang, Xuanjing},
  journal={arXiv preprint arXiv:2505.07591},
  year={2025}
}

@article{lexinstructeval2025,
  title={{LexInstructEval}: Lexical Instruction Following Evaluation for Large Language Models},
  author={Ren, Huimin and Liang, Yan and Su, Baiqiao and Sun, Chaobo and Lu, Hengtong and Zhang, Kaike and Wei, Chen},
  journal={arXiv preprint arXiv:2511.17561},
  year={2025}
}

@article{cctu2025,
  title={{CCTU}: A Benchmark for Tool Use under Complex Constraints},
  author={Ye, Junjie and Zhang, Guoqiang and Fu, Wenjie and Gui, Tao and Zhang, Qi and Huang, Xuanjing},
  journal={arXiv preprint arXiv:2603.15309},
  year={2026}
}

@article{ordermatters2025,
  title={Order Matters: Investigate the Position Bias in Multi-constraint Instruction Following},
  author={Zeng, Jie and He, Qianyu and Ren, Qingyu and Liang, Jiaqing and Xiao, Yanghua and Zhou, Weikang and Sun, Zeye and Yu, Fei},
  journal={arXiv preprint arXiv:2502.17204},
  year={2025}
}

@article{sun2024conifer,
  title={Conifer: Improving Complex Constrained Instruction-Following Ability of Large Language Models},
  author={Sun, Haoran and Liu, Lixin and Li, Junjie and Wang, Fengyu and Dong, Baohua and Lin, Ran and Huang, Ruohui},
  journal={arXiv preprint arXiv:2404.02823},
  year={2024}
}

@inproceedings{dong2025autoif,
  title={Self-Play with Execution Feedback: Improving Instruction-Following Capabilities of Large Language Models},
  author={Dong, Guanting and Lu, Keming and Li, Chengpeng and Xia, Tingyu and Yu, Bowen and Zhou, Chang and Zhou, Jingren},
  booktitle={The Thirteenth International Conference on Learning Representations},
  year={2025}
}

@book{mezard2009information,
  title={Information, Physics, and Computation},
  author={M{\'e}zard, Marc and Montanari, Andrea},
  year={2009},
  publisher={Oxford University Press}
}

@article{mertens2006threshold,
  title={Threshold Values of Random {K-SAT} from the Cavity Method},
  author={Mertens, Stephan and M{\'e}zard, Marc and Zecchina, Riccardo},
  journal={Random Structures \& Algorithms},
  volume={28},
  number={3},
  pages={340--373},
  year={2006}
}

@article{wei2022emergent,
  title={Emergent Abilities of Large Language Models},
  author={Wei, Jason and Tay, Yi and Bommasani, Rishi and Raffel, Colin and Zoph, Barret and Borgeaud, Sebastian and Yogatama, Dani and Bosma, Maarten and Zhou, Denny and Metzler, Donald and others},
  journal={Transactions on Machine Learning Research},
  year={2022}
}

@article{conceptmix2024,
  title={{ConceptMix}: A Compositional Image Generation Benchmark with Controllable Difficulty},
  author={Wu, Xindi and Yu, Dingli and Huang, Hangsibo and Russakovsky, Olga and Arora, Sanjeev},
  journal={arXiv preprint arXiv:2408.14339},
  year={2024}
}

@article{conceptmixpp2025,
  title={{ConceptMix++}: Leveling the Playing Field in Text-to-Image Benchmarking via Iterative Prompt Optimization},
  author={Gan, Haosheng and Tinaz, Berk and Sepehri, Mohammad Shahab and Fabian, Zalan and Soltanolkotabi, Mahdi},
  journal={arXiv preprint arXiv:2507.03275},
  year={2025}
}

@article{t2icompbench2023,
  title={{T2I-CompBench++}: An Enhanced and Comprehensive Benchmark for Compositional Text-to-image Generation},
  author={Huang, Kaiyi and Duan, Chengqi and Sun, Kaiyue and Xie, Enze and Li, Zhenguo and Liu, Xihui},
  journal={arXiv preprint arXiv:2307.06350},
  year={2023}
}

@article{t2vcompbench2024,
  title={{T2V-CompBench}: A Comprehensive Benchmark for Compositional Text-to-Video Generation},
  author={Sun, Kaiyue and Huang, Kaiyi and Liu, Xian and Wu, Yue and Xu, Zihan and Li, Zhenguo and Liu, Xihui},
  journal={arXiv preprint arXiv:2407.14505},
  year={2024}
}

@inproceedings{thrush2022winoground,
  title={{Winoground}: Probing Vision and Language Models for Visio-Linguistic Compositionality},
  author={Thrush, Tristan and Jiang, Ryan and Bartolo, Max and Singh, Amanpreet and Williams, Adina and Kiela, Douwe and Ross, Candace},
  booktitle={Proceedings of CVPR},
  year={2022}
}

@inproceedings{hsieh2023sugarcrepe,
  title={{SugarCrepe}: Fixing Hackable Benchmarks for Vision-Language Compositionality},
  author={Hsieh, Cheng-Yu and Zhang, Jieyu and Ma, Zixian and Kembhavi, Aniruddha and Krishna, Ranjay},
  booktitle={Advances in Neural Information Processing Systems},
volume = {36},
  year={2023}
}

@article{conme2024,
  title={{ConMe}: Rethinking Evaluation of Compositional Reasoning for Modern {VLMs}},
  author={Huang, Irene and Lin, Wei and Mirza, M. Jehanzeb and Hansen, Jacob A. and Doveh, Sivan and Butoi, Victor Ion and Herzig, Roei and Arbelle, Assaf and Kuehne, Hilde and Darrell, Trevor and Gan, Chuang and Oliva, Aude and Feris, Rogerio and Karlinsky, Leonid},
  journal={arXiv preprint arXiv:2406.08164},
  year={2024}
}

@article{genaibench2024,
  title={{GenAI-Bench}: Evaluating and Improving Compositional Text-to-Visual Generation},
  author={Li, Beiqi and Lin, Zhiqui and Pathak, Deepak and Li, Jiayao and Fei, Yixin and Wu, Kewen and Ling, Tiffany and Xia, Xide and Zhang, Pengchuan and Neubig, Graham and Ramanan, Deva},
  journal={arXiv preprint arXiv:2406.13743},
  year={2024}
}

@article{atlas2025,
  title={Adaptive Testing for {LLM} Evaluation: A Psychometric Alternative to Static Benchmarks},
  author={Li, Peiyu and Tang, Xiuxiu and Chen, Si and Cheng, Ying and Metoyer, Ronald and Hua, Ting and Chawla, Nitesh V.},
  journal={arXiv preprint arXiv:2511.04689},
  year={2025}
}

@article{adaptiveirt2025,
  title={Confident Rankings with Fewer Items: Adaptive {LLM} Evaluation with Continuous Scores},
  author={Balk{\i}r, Esma and Pernthaller, Alice and Basaldella, Marco and Hern{\'a}ndez-Orallo, Jos{\'e} and Collier, Nigel},
  journal={arXiv preprint arXiv:2601.13885},
  year={2026}
}

@article{impossiblebench2025,
  title={{ImpossibleBench}: Measuring {LLMs'} Propensity of Exploiting Test Cases},
  author={Zhong, Ziqian and Raghunathan, Aditi and Carlini, Nicholas},
  journal={arXiv preprint arXiv:2510.20270},
  year={2025}
}

@article{qi2024crab,
  title={Constraint Back-Translation Improves Complex Instruction Following of Large Language Models},
  author={Qi, Yunjia and Peng, Hao and Wang, Xiaozhi and Xu, Bin and Hou, Lei and Li, Juanzi},
  journal={arXiv preprint arXiv:2410.24175},
  year={2024}
}

@article{an2025ultraif,
  title={{ULTRA-IF}: Advancing Instruction Following from the Wild},
  author={An, Kaikai and Sheng, Li and Cui, Ganqu and Si, Shuzheng and Ding, Ning and Cheng, Yu and Chang, Baobao},
  journal={arXiv preprint arXiv:2502.04153},
  year={2025}
}

@article{liu2025air,
  title={{AIR}: Complex Instruction Generation via Automatic Iterative Refinement},
  author={Liu, Wei and He, Yancheng and Li, Yu and Huang, Hui and Hu, Chengwei and Liu, Jiaheng and Li, Shilong and Su, Wenbo and Zheng, Bo},
  journal={arXiv preprint arXiv:2502.17787},
  year={2025}
}

@inproceedings{cheng2025spar,
  title={{SPaR}: Self-Play with Tree-Search Refinement to Improve Instruction-Following in Large Language Models},
  author={Cheng, Jiale and Liu, Xiao and Wang, Cunxiang and Gu, Xiaotao and Lu, Yida and Zhang, Dan and Dong, Yuxiao and Tang, Jie and Wang, Hongning and Huang, Minlie},
  booktitle={The Thirteenth International Conference on Learning Representations},
  year={2025}
}

@article{chen2022controllable,
  title={Controllable Text Generation with Language Constraints},
  author={Chen, Howard and Li, Huihan and Chen, Danqi and Narasimhan, Karthik},
  journal={arXiv preprint arXiv:2212.10466},
  year={2022}
}

@inproceedings{ouyang2022training,
  title={Training Language Models to Follow Instructions with Human Feedback},
  author={Ouyang, Long and Wu, Jeffrey and Jiang, Xu and Almeida, Diogo and Wainwright, Carroll and Mishkin, Pamela and Zhang, Chong and Agarwal, Sandhini and Slama, Katarina and Ray, Alex and others},
  booktitle={Advances in Neural Information Processing Systems},
  year={2022}
}

@inproceedings{lu2021neurologic,
  title={{NeuroLogic} Decoding: (Un)supervised Neural Text Generation with Predicate Logic Constraints},
  author={Lu, Ximing and West, Peter and Zellers, Rowan and Le Bras, Ronan and Bhagavatula, Chandra and Choi, Yejin},
  booktitle={Proceedings of NAACL},
  year={2021}
}

@inproceedings{lu2022neurologic,
  title={{NeuroLogic A*esque} Decoding: Constrained Text Generation with Lookahead Heuristics},
  author={Lu, Ximing and Welleck, Sean and West, Peter and Jiang, Liwei and Kasai, Jungo and Khashabi, Daniel and Le Bras, Ronan and Qin, Lianhui and Yu, Youngjae and Zellers, Rowan and Smith, Noah A and Choi, Yejin},
  booktitle={Proceedings of NAACL},
  year={2022}
}

@inproceedings{ferraz2024decrim,
  title={LLM Self-Correction with {DeCRIM}: Decompose, Critique, and Refine for Enhanced Following of Instructions with Multiple Constraints},
  author={Ferraz, Thomas Palmeira and Mehta, Kartik and Lin, Yu-Hsiang
          and Chang, Haw-Shiuan and Oraby, Shereen and Liu, Sijia
          and Subramanian, Vivek and Chung, Tagyoung and Bansal, Mohit
          and Peng, Nanyun},
  booktitle={Advances in Neural Information Processing Systems},
  year={2024}
}

@article{lu2025miso,
  title={Enhancing Complex Instruction Following for Large Language Models
         with Mixture-of-Contexts Fine-tuning},
  author={Lu, Yuheng and Bai, ZiMeng and Yuan, Caixia
          and Jiang, Huixing and Wang, Xiaojie},
  journal={arXiv preprint arXiv:2505.11922},
  year={2025}
}

@inproceedings{jiang2024complex2simple,
  title={From Complex to Simple: Enhancing Multi-Constraint Complex
         Instruction Following Ability of Large Language Models},
  author={Jiang, Qianyu and Zeng, Jie and He, Qianxi and Liang, Jiaqing and Xiao, Yanghua},
  booktitle={Findings of the Association for Computational Linguistics: EMNLP},
  year={2024}
}

@article{kaplan2020scaling,
  title={Scaling Laws for Neural Language Models},
  author={Kaplan, Jared and McCandlish, Sam and Henighan, Tom and Brown, Tom B and Chess, Benjamin and Child, Rewon and Gray, Scott and Radford, Alec and Wu, Jeffrey and Amodei, Dario},
  journal={arXiv preprint arXiv:2001.08361},
  year={2020}
}

@inproceedings{hoffmann2022chinchilla,
  title={Training Compute-Optimal Large Language Models},
  author={Hoffmann, Jordan and Borgeaud, Sebastian and Mensch, Arthur and Buchatskaya, Elena and Cai, Trevor and Rutherford, Eliza and Casas, Diego de Las and Hendricks, Lisa Anne and Welbl, Johannes and Clark, Aidan and others},
  booktitle={Advances in Neural Information Processing Systems},
  volume={35},
  pages={30016--30030},
  year={2022}
}

@article{cohen2024compositional,
  title={Compositional Instruction Following with Language Models and Reinforcement Learning},
  author={Cohen, Vanya and Nangue Tasse, Geraud and Gopalan, Nakul and James, Steven and Gombolay, Matthew and Mooney, Raymond and Rosman, Benjamin},
  journal={Transactions on Machine Learning Research},
  year={2024},
  note={arXiv:2501.12539}
}

@article{sakai2025ordered,
  title={Revisiting Compositional Generalization Capability of Large Language Models Considering Instruction Following Ability},
  author={Sakai, Yusuke and Kamigaito, Hidetaka and Watanabe, Taro},
  journal={arXiv preprint arXiv:2506.15629},
  year={2025}
}

@inproceedings{lv2026ata,
  title={Attend to the Active: Structure-Aware Dynamic Attention in {LLMs} for Compositional Instruction Following},
  author={Lv, Fangrui and Qin, Yulei and Hong, Ruixin and Liang, Jian and Wu, Jinyang and Li, Ke and Sun, Xing and Zhang, Changshui},
  booktitle={International Conference on Learning Representations (ICLR)},
  year={2026}
}

@inproceedings{favero2025compositional,
  title={How Compositional Generalization and Creativity Improve as Diffusion Models are Trained},
  author={Favero, Alessandro and Sclocchi, Antonio and Cagnetta, Francesco and Frossard, Pascal and Wyart, Matthieu},
  booktitle={Forty-second International Conference on Machine Learning},
  year={2025}
}

@inproceedings{dziri2023faith,
  title={Faith and Fate: Limits of Transformers on Compositionality},
  author={Dziri, Nouha and Lu, Ximing and Sclar, Melanie and Li, Xiang Lorraine and Jiang, Liwei Jiang and Lin, Bill Yuchen and West, Peter and Bhagavatula, Chandra and Le Bras, Ronan and Hwang, Jena D and Sanyal, Soumya and Welleck, Sean and Ren, Xiang and Ettinger, Allyson and Harchaoui, Zaid and Choi, Yejin},
  booktitle={Advances in Neural Information Processing Systems},
  volume={36},
  year={2023}
}

@inproceedings{press2023measuring,
  title={Measuring and Narrowing the Compositionality Gap in Language Models},
  author={Press, Ofir and Zhang, Muru and Min, Sewon and Schmidt, Ludwig and Smith, Noah A and Lewis, Mike},
  booktitle={Findings of the Association for Computational Linguistics: EMNLP},
  year={2023}
}

@inproceedings{lake2018generalization,
  title={Generalization without Systematicity: On the Compositional Skills of Sequence-to-Sequence Recurrent Networks},
  author={Lake, Brenden M and Baroni, Marco},
  booktitle={International Conference on Machine Learning (ICML)},
  year={2018}
}

@inproceedings{beurer2023lmql,
  title={Prompting Is Programming: A Query Language for Large Language Models},
  author={Beurer-Kellner, Luca and Fischer, Marc and Vechev, Martin},
  booktitle={Proceedings of the 44th ACM SIGPLAN International Conference on Programming Language Design and Implementation (PLDI)},
  year={2023},
  doi={10.1145/3591300}
}

@article{cowan2001magical,
  title={The Magical Number 4 in Short-Term Memory: A Reconsideration of Mental Storage Capacity},
  author={Cowan, Nelson},
  journal={Behavioral and Brain Sciences},
  volume={24},
  number={1},
  pages={87--114},
  year={2001},
  publisher={Cambridge University Press},
  doi={10.1017/S0140525X01003922}
}

@inproceedings{liang2025figv,
  title={Fine-Grained Constraint Generation-Verification for Improved Instruction-Following},
  author={Liang, Zhixiang and Hou, Zhenyu and Wang, Xiao},
  booktitle={Proceedings of the Fourth Workshop on Generation, Evaluation and Metrics (GEM\textsuperscript{2})},
  pages={862--879},
  year={2025},
  address={Vienna, Austria},
  publisher={Association for Computational Linguistics}
}

@article{miller1956magical,
  title={The Magical Number Seven, Plus or Minus Two: Some Limits on Our Capacity for Processing Information},
  author={Miller, George A.},
  journal={Psychological Review},
  volume={63},
  number={2},
  pages={81--97},
  year={1956},
  doi={10.1037/h0043158}
}

@article{willard2023outlines,
  title={Efficient Guided Generation for Large Language Models},
  author={Willard, Brandon T. and Louf, R{\'e}mi},
  journal={arXiv preprint arXiv:2307.09702},
  year={2023},
  doi={10.48550/arXiv.2307.09702}
}

\appendix
\section{Reproducibility}
\label{app:reproducibility}

All probes, deterministic verifiers, raw model outputs, and evaluation code will be released upon publication.
The total evaluation comprises 67{,}905 API calls (15 models $\times$ 4{,}527 probes) at temperature${=}0$ with greedy decoding.
Four of 15 evaluated models are open-weight and fully reproducible: Llama~3.1~70B and 405B Instruct, Llama~4~Maverick, and Qwen~3~235B Instruct.
Proprietary model outputs are included in the release for verification.
The probe composer, compatibility checker, calibrated parameter sampling distributions, and all 36 verification functions are included in the released code.

\section{Related Work (Expanded)}
\label{app:related_full}

This section provides an expanded version of the related work discussion summarized in \cref{sec:intro}.

\subsection{Compositional Instruction Following}

\paragraph{Evaluation benchmarks.}
IFEval \cite{zhou2023instruction} established the paradigm of verifiable instruction following with 25 constraint types and deterministic programmatic checks, but probes typically contain $k \leq 3$ constraints---insufficient to observe compositional degradation.
A model scoring 80\%+ on IFEval could in principle collapse at $k{=}5{-}8$ on CSE, exposing degradation invisible to proficiency-oriented evaluation.
Adaptive IRT methods \cite{adaptiveirt2025} extend the evaluation paradigm by enabling confident model rankings with fewer test items, though they require more examinees than CSE's 15-model panel.
InfoBench \cite{qin2024infobench} decomposes each instruction into atomic verifiable criteria and evaluates per-criterion satisfaction via its Decomposed Requirements Following Rate (DRFR). GPT-4 achieves ${\sim}89\%$ DRFR overall, yet more than 10\% of requirements remain unfulfilled even for this strongest model, and no model fully satisfies number or linguistic constraints. Performance is highest on content and style, intermediate on format, and lowest on number and linguistic dimensions---a constraint-type hierarchy that foreshadows CSE's depth-of-processing finding, where structural constraints degrade $2.0{\times}$ faster than lexical ones. InfoBench operates at the single-instruction level; CSE extends this analysis systematically across $k{=}1$ to $12$, revealing how per-criterion gaps compound into the phase transition.
FollowBench \cite{jiang2024followbench} scales to $k{=}5$ and introduces the Consistent Satisfaction Level (CSL) metric for tracking partial compliance across difficulty levels, but relies on GPT-4 as a judge for several constraint types, introducing evaluation variance.
CSL measures which satisfaction level a model can sustain, which is conceptually adjacent to CSE's $k$-curve---but CSE isolates where the transition occurs rather than tracking cumulative levels, and uses fully deterministic verification throughout.
COLLIE \cite{yao2023collie} is the closest prior work: it pioneered grammar-based constraint specification with formal CFG definitions and corpus-grounded extraction ensuring natural solutions exist, and demonstrated that compositional constraints are harder than individual ones---a finding we confirm and extend.
Three key differences: (1)~COLLIE uses fixed constraint compositions while CSE systematically varies $k$ from 1--12, enabling the decay characterization that is this paper's central contribution; (2)~COLLIE reports that composition is hard but does not quantify \emph{where} capability breaks or what functional form the degradation takes; (3)~CSE's dual scoring (binary + continuous) reveals gaps between partial compliance and strict satisfaction that are absent from COLLIE's binary-only evaluation.
COLLIE is stronger in formal rigor (CFG grammar) and in its pass@$k$ evaluation design; CSE contributes the systematic scaling analysis and the mechanistic decomposition of why composition fails.

\paragraph{Complex constraint benchmarks.}
RECAST \cite{guo2025recast} pushes constraint counts to $k{=}13{+}$ with 19 constraint types and proposes RLVC training with constraint-specific reward signals, but does not report per-$k$ performance trajectories, precluding decay analysis.
ComplexBench \cite{complexbench2024} models constraint composition \emph{structure}---And, Chain, Selection, Nesting---rather than count, and is thus orthogonal to our analysis of how many constraints can be jointly satisfied.
IFBENCH \cite{ifbench2025} demonstrates that models overfit to IFEval's 25 known constraint types and proposes IF-RLVR training for generalization---a different axis (unseen constraints) from our focus (composition scaling).
CCTU \cite{cctu2025} evaluates constrained tool use, finding that no model exceeds 20\% task completion under strict multi-constraint adherence---a ceiling remarkably close to CSE's probe-level floor at $k{\geq}9$, despite a completely different task domain, suggesting the phase transition may reflect a general property of LLM compositional processing.

\paragraph{Training methods.}
Conifer \cite{sun2024conifer} generates multi-level constraints with GPT-4 and employs curriculum tuning, while AutoIF \cite{dong2025autoif} implements self-dialogue with execution-based verification; AIR \cite{liu2025air} and MiSO \cite{lu2025miso} extend automatic refinement and mixture-of-contexts fine-tuning, respectively, to complex instruction generation---all focus on training rather than diagnostic evaluation.
MulDimIF \cite{ye2025muldimif} explores how constraint \emph{presentation mode} (example, listing, incorporation) affects following and proposes GRPO training; LexInstructEval \cite{lexinstructeval2025} introduces a formal grammar for lexical constraints with bilingual evaluation, finding that procedural depth degrades performance---complementary to our finding that constraint \emph{count} does so.
UltraIF \cite{an2025ultraif} scales to 100K+ instructions via automatic constraint decomposition; SPaR \cite{cheng2025spar} uses self-play with tree-search refinement for similar scaling.
DeCRIM \cite{ferraz2024decrim} takes a complementary approach, decomposing multi-constraint instructions into individual constraints and iteratively refining compliance; related decomposition strategies include Complex-to-Simple \cite{jiang2024complex2simple} and constraint back-translation \cite{qi2024crab}---approaches whose effectiveness CSE's independence finding predicts, since failures accumulate per-constraint rather than through pairwise interference.
FiGV \cite{liang2025figv} synthesizes constraint-augmented training data with hybrid LLM/function verification at the constraint level, confirming that deterministic verification of individual constraints produces higher-quality training signal than holistic LLM judgment---a principle CSE applies to evaluation.
\citet{chen2022controllable} formalize controllable generation as satisfaction of language constraints, bridging constrained decoding and instruction following.
Neurologic Decoding \cite{lu2021neurologic,lu2022neurologic} approaches constraint satisfaction from the inference side via lexically-constrained beam search; more recent structured generation frameworks \cite{willard2023outlines,beurer2023lmql} can enforce format and keyword constraints at the decoding level, removing them from the model's compositional burden entirely.
CSE evaluates raw generation without such scaffolding to isolate intrinsic compositional capacity; measuring how constrained decoding or chain-of-thought prompting shifts the transition is a natural follow-up (\cref{sec:discussion}).

\paragraph{Mechanistic explanations.}
ATA \cite{lv2026ata} identifies a mechanistic explanation for compositional degradation: models diffusely attend across all sub-tasks in the input, and inactive sub-tasks cause interference via attention distraction---corroborating our phase transition with performance drops from 60\% to 48\% as sub-task count increases.
Their inference-time attention steering recovers 6--10pp without retraining, demonstrating that the transition is partially addressable through architectural intervention.
In reinforcement learning, CERLLA \cite{cohen2024compositional} demonstrates that compositional value function representations with Boolean task algebra (AND, OR, NOT) enable sample-efficient learning of 162 language-conditioned tasks.
CSE's constraint composition is purely conjunctive (all $k$ constraints must hold simultaneously), but the shared insight is that compositional structure---whether in policy space or constraint space---determines scalability.

\subsection{Constraint Scaling and Cross-Modal Parallels}

\paragraph{Interaction vs.\ accumulation.}
A central question in compositional evaluation is whether difficulty arises from constraint \emph{interactions} (specific pairs that conflict) or from \emph{accumulation} (each additional constraint independently reducing success probability).
ComplexBench \cite{complexbench2024} models structural interactions through nesting and branching but does not measure pairwise statistical independence.
\citet{ordermatters2025} demonstrate that constraint ordering affects performance by up to 25\%, establishing that presentation matters---we find this effect is orthogonal to constraint count (\cref{sec:cofailure}).
Ordered CommonGen \cite{sakai2025ordered} evaluates whether LLMs can generate sentences with concepts in a specified order across 36 models, finding that even the best model follows the specified order only 75\% of the time on a single ordering constraint.
CSE's ordering constraints show similarly low $k{=}1$ rates (O1 monotonic length: 28.9\%, O2 alphabetical: 43.7\%), and these per-constraint imperfections are precisely what compound multiplicatively to produce the phase transition.

\paragraph{Phase transitions and scaling laws.}
Our finding that constraint failures are weakly correlated connects to the theoretical literature on phase transitions in constraint satisfaction.
Random $k$-SAT exhibits a sharp satisfiability threshold driven by clause density and interaction structure \cite{mertens2006threshold,mezard2009information}.
Our transition is mechanistically simpler: it arises from near-multiplicative accumulation of failure probabilities without requiring interaction structure.
This parallels neural scaling laws \cite{kaplan2020scaling,hoffmann2022chinchilla}, which established that simple parametric models (power laws in compute and data) predict performance with practical accuracy.
\citet{favero2025compositional} provide theoretical support for hierarchical compositional learning: diffusion models learn grammar rules level-by-level, with sample complexity $P_\ell \propto m^{\ell+1}$ for level-$\ell$ rules---higher-level (more global) rules require polynomially more data.
This predicts that constraints requiring deeper compositional processing should be more fragile, consistent with our finding that structural constraints degrade $2.0{\times}$ faster than lexical constraints under compositional load (\cref{sec:degradation}).
Our floor model $P(k) = a \cdot e^{-\beta k} + c$ serves an analogous role for constraint composition: three parameters, fitted on $k{=}1{-}8$, predict held-out $k{=}9{-}12$ within 0.2pp---comparable predictive precision from a comparably simple functional form.
The ``emergent abilities'' literature \cite{wei2022emergent} documents sharp capability transitions as a function of model scale; we find analogous transitions as a function of compositional load.

\paragraph{Cross-modal compositional decay.}
Earlier work on vision-language compositionality \cite{thrush2022winoground,hsieh2023sugarcrepe} establishes that VLMs struggle with compositional attribute binding even at $k{=}2$; \citet{conme2024} show that these failures persist under controlled evaluation designs.
The decay pattern we observe in text is consistent with cross-modal findings in generation tasks.
ConceptMix \cite{conceptmix2024} reports that text-to-image models drop from 83\% at $k{=}1$ concepts to 8\% at $k{=}7$---implying $r \approx 0.65$ per added concept, steeper than CSE's aggregate $r{=}0.922$ but following the same multiplicative functional form.
ConceptMix++ \cite{conceptmixpp2025} shows that prompt optimization can recover up to 20\% of lost compositional capability, suggesting the limitation is partially in prompt understanding rather than solely in generation.
T2I-CompBench++ \cite{t2icompbench2023} documents degradation across attribute binding, spatial relationships, and generative numeracy.
T2V-CompBench \cite{t2vcompbench2024} shows temporal composition failures in video generation.
GenAI-Bench \cite{genaibench2024} evaluates compositional text-to-visual generation, reporting degradation with compositional complexity in image, video, and 3D generation.
These are all constrained \emph{generation} tasks, and the consistency of the decay pattern across text-to-text, text-to-image, and text-to-video generation suggests the transition may be a general property of compositional generation rather than a modality-specific limitation.
ImpossibleBench \cite{impossiblebench2025} tests whether models detect infeasible constraint combinations in coding tasks; CSE includes deliberate impossible probes and finds zero hallucinated compliance---models either correctly refuse or fail on the constraint text itself.

\section{Per-Constraint mCSR Heatmap}
\label{app:heatmaps}

\begin{figure*}[h]
\centering
\includegraphics[width=\textwidth, trim={30pt 0pt 100pt 0pt}, clip]{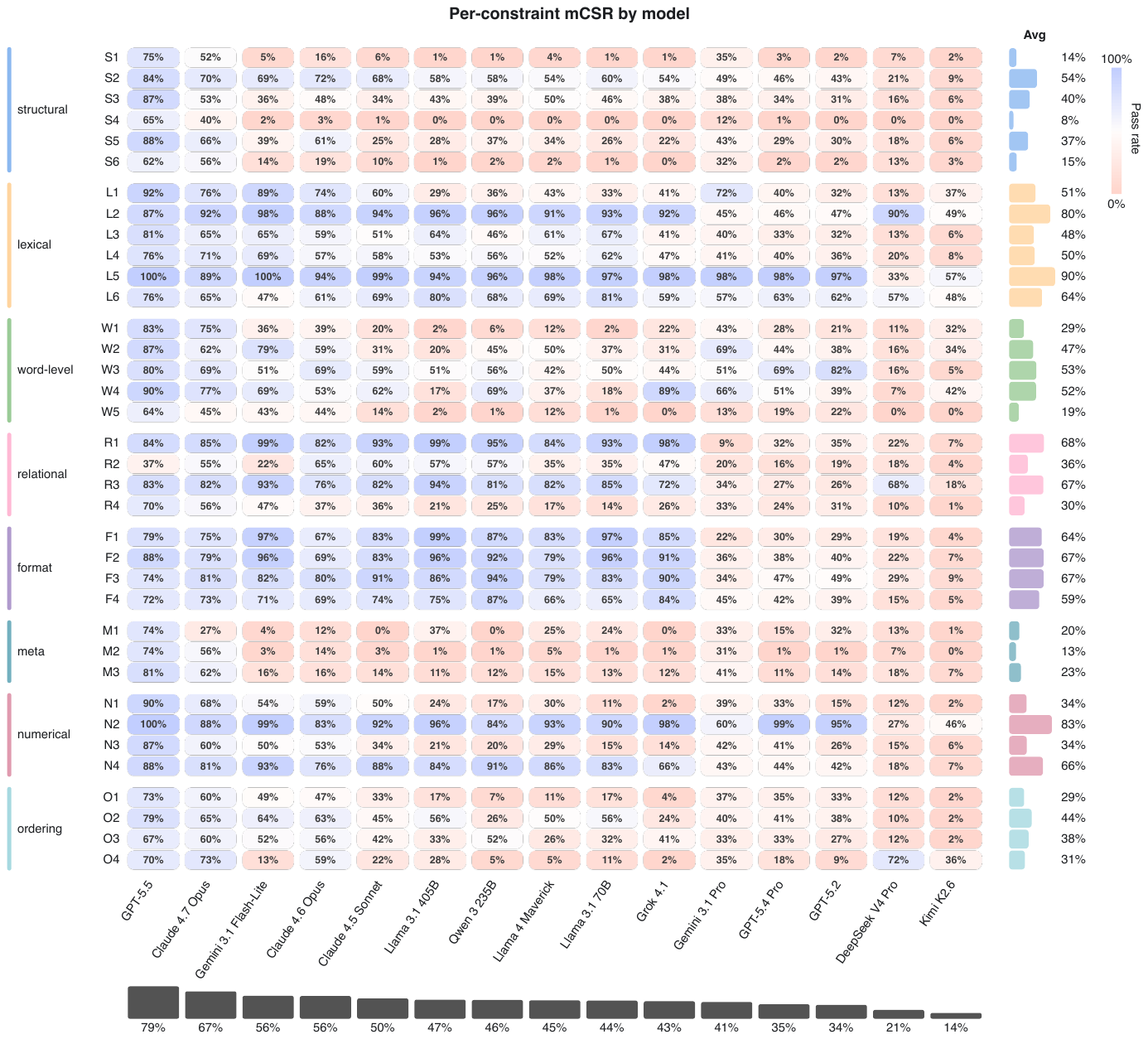}
\caption{Per-constraint mCSR by model (36 constraints $\times$ 15 models). GPT-5.5 dominates across all dimensions (79.8\% overall), followed by Claude~4.7~Opus (67.0\%) and Flash-Lite (56.6\%). The format--relational cluster is visible as a correlated block (F1--F3, R1, R3): models that fail structured output fail relational reasoning simultaneously---output-feature coupling, not cognitive interference (\cref{sec:cofailure}). Structural and meta constraints (S1, S4, S6, M2) are uniformly hard (${\leq}15\%$), while lexical constraints (L5 forbidden word: 90\%, L2 mandatory words: 80\%) and numerical (N2 no digits: 83\%) sustain the asymptotic floor. DeepSeek~V4~Pro and Kimi~K2.6 show the fast-decay pattern: high variance across constraints but uniformly low rates outside the immune cluster.}
\label{fig:cse_e_heatmap}
\end{figure*}

\Cref{fig:cse_e_heatmap} shows per-constraint mCSR for all 36 constraints across 15 models.

\section{Constraint Profiles}
\label{app:constraint_genome}

\begin{figure*}[h]
\centering
\includegraphics[width=\textwidth, trim={10pt 10pt 10pt 10pt}, clip]{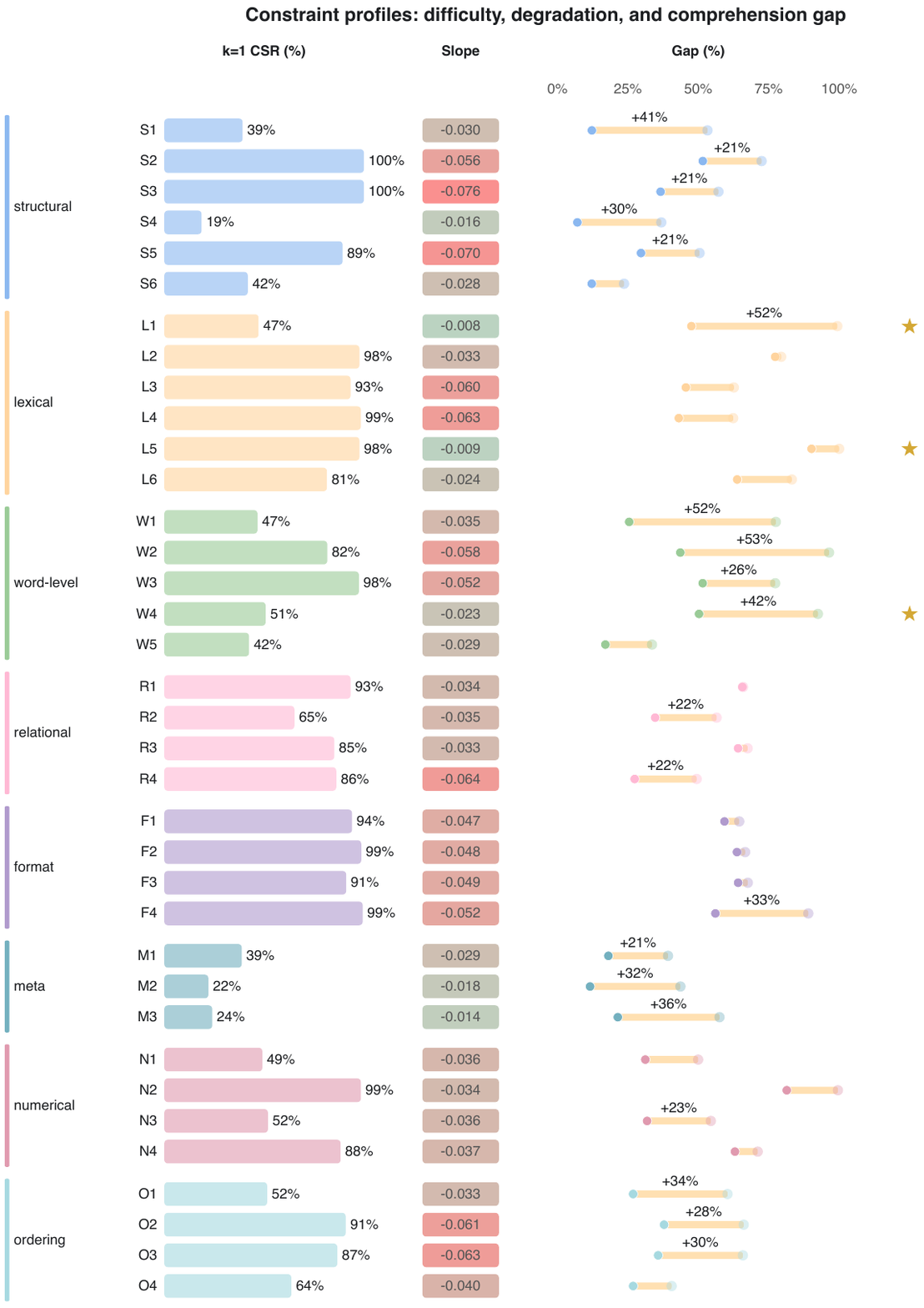}
\caption{Constraint profiles for all 36 constraints, grouped by dimension. \textbf{Left:} $k{=}1$ mCSR (single-constraint difficulty in isolation).
  \textbf{Center:} per-constraint degradation slope (more negative = faster collapse under compositional load; warm = fast, cool = slow). \textbf{Right:}
  comprehension-maintenance gap $\Delta_i = \bar{s}_i - \text{mCSR}_i$ (\cref{eq:gap}): filled circles show mCSR (strict binary satisfaction), open circles show continuous score (partial compliance), and the connecting bar is $\Delta$. Constraints with $\Delta > 20$pp are labeled. A large gap indicates that models partially comply---demonstrating comprehension---but fail strict verification (e.g., W1 unique words: score 78.6\%, mCSR 28.4\%, $\Delta{=}50$pp). A near-zero gap indicates all-or-nothing behavior with no partial-credit regime (e.g., R1 scene graph: score 67.9\%, mCSR 67.7\%, $\Delta{=}0.2$pp). Stars mark compositionally immune constraints (retention ${>}80\%$ at $k{\geq}8$): L1, L5, W4.}
\label{fig:constraint_genome}
\end{figure*}

\Cref{fig:constraint_genome} visualizes three properties of each constraint side by side: isolated difficulty ($k{=}1$ mCSR), compositional vulnerability (degradation slope), and the comprehension-maintenance gap.

\section{Complete Phase Transition Table}
\label{app:phase_transition}

\Cref{tab:phase_transition} reports per-model mCSR at each $k$ for all 15 models, sorted by $k{=}1$ performance. GPT-5.5 leads at every $k$ value; the per-model decay factors $r$ (\cref{sec:decay_model}) compress each trajectory into a single number.

\definecolor{slowdecay}{HTML}{D6F5D6}
\definecolor{meddecay}{HTML}{FFF0D6}
\definecolor{fastdecay}{HTML}{FFD6D6}

\begin{table*}[h]
\caption{Per-model mCSR (\%) by constraint count $k$, sorted by $k{=}1$. \textbf{Bold} marks the best model at each $k$. The rightmost column $r$ is the per-model decay factor from $\text{rate}(k) = a \cdot r^{k-1}$, fitted on $k{=}1{-}8$: the fraction of performance retained per additional constraint ($r{=}0.95$ means 5\% loss per constraint). All 15~models start between 48--98\% at $k{=}1$ and converge toward a model-dependent floor at $k{\geq}8$. $r$ ranges from $0.721$ (DeepSeek, loses 28\% per constraint) to $0.977$ (Claude~4.7, loses 2\%). Row shading indicates decay archetype: unshaded is slow decay ($r{>}0.93$, 10 models), \colorbox{meddecay}{medium} ($0.80{\leq}r{\leq}0.93$), \colorbox{fastdecay}{fast} ($r{<}0.80$).}
\label{tab:phase_transition}
\begin{center}
\begin{small}
\begin{tabular}{l*{12}{r}r}
\toprule
\textbf{Model} & $k{=}1$ & $2$ & $3$ & $4$ & $5$ & $6$ & $7$ & $8$ & $9$ & $10$ & $11$ & $12$ & $r$ \\
\midrule
GPT-5.5              & \textbf{98.4} & \textbf{96.0} & \textbf{92.9} & \textbf{90.5} & \textbf{90.4} & \textbf{85.8} & \textbf{83.8} & \textbf{80.2} & \textbf{73.4} & \textbf{68.2} & \textbf{65.1} & \textbf{53.0} & .973 \\
\rowcolor{meddecay} Gemini 3.1 Pro       & 90.5 & 82.5 & 71.7 & 59.6 & 51.0 & 45.9 & 38.7 & 30.2 & 26.6 & 21.8 & 21.5 & 17.9 & .863 \\
\rowcolor{fastdecay} DeepSeek V4 Pro      & 80.8 & 71.2 & 45.9 & 28.0 & 18.1 & 13.7 & 14.4 & 12.1 & 11.9 & 12.5 & 13.0 & 11.6 & .721 \\
Claude 4.6 Opus      & 80.3 & 80.2 & 77.1 & 73.3 & 70.1 & 65.0 & 60.0 & 50.2 & 45.7 & 37.3 & 32.6 & 23.5 & .944 \\
Claude 4.7 Opus      & 78.9 & 83.3 & 84.8 & 83.5 & 80.2 & 78.0 & 73.0 & 66.3 & 54.0 & 44.0 & 26.5 & 37.2 & \textbf{.977} \\
Flash-Lite           & 78.1 & 76.2 & 68.9 & 64.6 & 61.3 & 59.1 & 53.6 & 53.8 & 50.6 & 47.9 & 47.5 & 42.4 & .943 \\
\rowcolor{meddecay} GPT-5.4 Pro          & 77.6 & 72.2 & 66.3 & 55.5 & 49.7 & 40.1 & 26.3 & 24.5 & 17.3 & 15.4 & 15.2 & 14.4 & .859 \\
\rowcolor{meddecay} GPT-5.2              & 68.4 & 65.3 & 61.8 & 51.2 & 47.2 & 36.1 & 26.6 & 23.3 & 18.7 & 16.2 & 15.1 & 15.3 & .869 \\
Llama 4 Maverick     & 65.1 & 61.9 & 55.5 & 51.1 & 47.0 & 44.8 & 44.3 & 42.7 & 40.1 & 39.2 & 37.9 & 36.3 & .935 \\
Claude 4.5 Sonnet    & 64.3 & 68.8 & 62.8 & 59.3 & 56.2 & 53.5 & 50.8 & 45.9 & 42.6 & 39.9 & 36.0 & 31.7 & .951 \\
Llama 3.1 405B       & 62.4 & 61.5 & 54.6 & 52.3 & 49.1 & 47.3 & 46.4 & 43.0 & 41.9 & 41.1 & 39.4 & 37.3 & .947 \\
Qwen 3 235B Inst.    & 56.8 & 61.3 & 54.9 & 50.0 & 48.9 & 47.6 & 45.4 & 44.0 & 40.0 & 40.1 & 38.2 & 35.8 & .956 \\
Llama 3.1 70B        & 56.2 & 56.5 & 52.7 & 51.0 & 46.3 & 44.8 & 44.0 & 41.2 & 40.9 & 40.0 & 38.6 & 36.3 & .953 \\
Grok 4.1             & 54.9 & 58.3 & 50.3 & 48.5 & 42.7 & 44.3 & 40.9 & 40.3 & 38.4 & 37.6 & 38.8 & 34.3 & .948 \\
\rowcolor{fastdecay} Kimi K2.6            & 48.4 & 29.8 & 20.3 & 16.9 & 12.3 & 13.1 & 13.3 & 14.7 & 12.0 & 12.5 & 13.2 & 11.8 & .769 \\
\bottomrule
\end{tabular}
\end{small}
\end{center}
\end{table*}

\section{Decay Model Predictions and Cross-Validation}
\label{app:decay_detail}

\definecolor{traincolor}{HTML}{D6F5D6}
\definecolor{testcolor}{HTML}{FFF0D6}

\paragraph{Aggregate decay model.}
The per-constraint aggregate decay is well-described by a pure multiplicative model: $\text{rate}(k) = 72.0\% \times 0.922^{k-1}$ (held-out MAE~$= 0.2$pp on $k{=}9{-}12$).
A three-parameter floor model ($P(k) = a \cdot e^{-\beta k} + c$, all-data fit: $a{=}0.782$, $\beta{=}0.082$, $c{=}0.0\%$) does not improve on this aggregate fit (MAE~$= 0.2$pp), because averaging across 15~models with diverse decay profiles ($r = 0.72{-}0.98$) smooths the per-model floors into a continuous decline.

\paragraph{Cross-validation.}
The floor parameter $c$ is unstable under cross-validation: it collapses to 0.0\% in 11 of 12 LOO folds and inflates to 9.4\% when $k{=}1$ is held out, indicating that 12~data points do not robustly constrain a three-parameter model at the aggregate level.
Under multiple train/test splits, the floor model performs comparably to, but does not outperform, the multiplicative baseline.

\paragraph{The floor is per-model, not aggregate.}
The aggregate curve masks per-model floors because between-model variance in decay rate ($r = 0.72{-}0.98$) dominates the signal.
At $k{=}12$, the multiplicative model systematically underpredicts 10 of 15~models (mean error: $-5.1$pp), indicating that these models plateau above the multiplicative prediction.
Even the weakest models (DeepSeek~V4~Pro, Kimi~K2.6) sustain ${\sim}11\%$ per-constraint rates at $k{=}12$ despite multiplicative predictions of $1{-}3\%$---the compositionally immune constraints keep them off zero.
Two models (Claude~4.6~Opus, Claude~4.7~Opus) are overpredicted, decaying \emph{faster} than the multiplicative model at high~$k$.
The floor is a per-model phenomenon, visible when models are analyzed individually but masked in the cross-model aggregate.

\paragraph{Independent floor derivation.}
Compositionally immune constraints are identified by two criteria: degradation slope $> {-}0.02$ (nearly flat across $k$) and $k{\geq}8$ pass rate $> 5\%$ (non-trivially contributing to the floor).
Four constraints meet both criteria: L5~(forbidden word, 87.4\% at $k{\geq}8$), L1~(lipogram, 45.0\%), M3~(word count divisibility, 18.3\%), and M2~(self-counting, 7.6\%).
S4 meets the slope criterion but has a trivially low rate (4.5\% at $k{\geq}8$).
W4~(no repeated bigrams) narrowly misses the slope threshold ($-0.021$) but sustains 43.7\% at $k{\geq}8$ (86\% retention), qualifying as immune under the retention criterion used in \cref{sec:degradation}.
The independently predicted floor from these immune constraints is $c{=}34.5\%$ (mean $k{\geq}8$ rate of the four immune constraints, weighted by their representation)---far above the fitted $c{=}0.0\%$.
The gap confirms that the immune constraints' per-constraint rates do not translate directly to an aggregate floor: their contribution is diluted by the 32~non-immune constraints whose rates converge toward zero, and by the between-model variance that smooths the aggregate curve.

\paragraph{Floor estimate comparison.}
\begin{center}
\begin{small}
\begin{tabular}{lr}
\toprule
\textbf{Method} & $c$ \\
\midrule
All-data fit ($k{=}1{-}12$) & 0.0\% \\
LOO mean $\pm$ std & 0.8\% $\pm$ 2.7\% \\
$k{=}1{-}8$ extrapolation & 0.0\% \\
Independent (immune constraints) & 34.5\% \\
\bottomrule
\end{tabular}
\end{small}
\end{center}
The fitted and LOO estimates converge near zero; the independent estimate overshoots because it counts immune constraint rates without accounting for their dilution in the 36-constraint aggregate.
The floor is real per-constraint (L5 at 87\%, L1 at 45\%, W4 at 44\%, M3 at 18\% at $k{\geq}8$) and per-model (10/15 underpredicted at $k{=}12$), but is not a feature of the aggregate decay curve with 15~models.

\section{Failure Mode Details}
\label{app:l1_failure}

This appendix provides per-model breakdowns supporting the failure mode analysis in \cref{sec:degradation}.

\begin{table}[h]
\caption{L1 (lipogram) failure categorization by model ($n{=}5{,}267$ total failures across 15~models). ``Genuine'' = model attempts the constraint and violates it in content; ``Refusal'' = model identifies infeasibility and the forbidden letter appears in the explanation text; ``Meta'' = constraint self-narration (restating the rule using the forbidden letter); ``Case'' = case-folding artifact. The refusal rate varies widely: GPT-5.2 and GPT-5.4 refuse ${\sim}90\%$ of failed probes, while open-weight models (Llama, Grok, Maverick) refuse ${<}1\%$.}
\begin{center}
\begin{small}
\setlength{\tabcolsep}{2pt}
\begin{tabular}{@{}lrrrrr@{}}
\toprule
\textbf{Model} & \textbf{Genuine} & \textbf{Meta} & \textbf{Refusal} & \textbf{Case} & \textbf{Total} \\
\midrule
Claude 4.5 Sonnet & 235 & 52 & 1 & 0 & 288 \\
Claude 4.6 Opus & 156 & 32 & 1 & 0 & 189 \\
Claude 4.7 Opus & 72 & 46 & 28 & 0 & 146 \\
DeepSeek V4 Pro & \textbf{566} & 30 & 3 & 29 & 628 \\
Gemini 3.1 Flash-Lite & 76 & 1 & 0 & 0 & 77 \\
Gemini 3.1 Pro & 144 & 3 & 56 & 1 & 204 \\
GPT-5.2 & 50 & 0 & \textbf{443} & 0 & 493 \\
GPT-5.4 Pro & 34 & 0 & \textbf{397} & 0 & 431 \\
GPT-5.5 & 8 & 0 & 50 & 0 & 58 \\
Grok 4.1 & 427 & 0 & 0 & 0 & 427 \\
Kimi K2.6 & 394 & 46 & 0 & 16 & 456 \\
Llama 3.1 405B & 511 & 0 & 0 & 0 & 511 \\
Llama 3.1 70B & 485 & 0 & 0 & 0 & 485 \\
Llama 4 Maverick & 409 & 0 & 0 & 0 & 409 \\
Qwen 3 235B Inst. & 455 & 0 & 10 & 0 & 465 \\
\midrule
\textbf{Total} & \textbf{4{,}022} & \textbf{210} & \textbf{989} & \textbf{46} & \textbf{5{,}267} \\
\bottomrule
\end{tabular}
\end{small}
\end{center}
\end{table}

\definecolor{immcolor}{HTML}{FFD6D6}
\definecolor{gradcolor}{HTML}{D6E8FF}

\begin{table}[h]
\caption{Model failure strategy profiles for genuine L1 failures only ($n{=}4{,}022$). Median = normalized position of first violation (0 = response start, 1 = end). \colorbox{immcolor}{Pink} = immediate strategy (median $< 0.05$): violations occur at response onset. \colorbox{gradcolor}{Blue} = gradual strategy (median ${\geq}0.10$): violations distributed across the response. GPT-5.5 is excluded (8 genuine failures, insufficient for profiling).}
\begin{center}
\begin{small}
\begin{tabular}{lrr}
\toprule
\textbf{Model} & \textbf{Median} & $n$ \\
\midrule
\rowcolor{immcolor} DeepSeek V4 Pro & 0.010 & 566 \\
\rowcolor{immcolor} Claude 4.7 Opus & 0.012 & 72 \\
\rowcolor{immcolor} Claude 4.6 Opus & 0.028 & 156 \\
\rowcolor{immcolor} Kimi K2.6 & 0.029 & 394 \\
\rowcolor{immcolor} Claude 4.5 Sonnet & 0.048 & 235 \\
\midrule
\rowcolor{gradcolor} Llama 4 Maverick & 0.105 & 409 \\
\rowcolor{gradcolor} Llama 3.1 70B & 0.115 & 485 \\
\rowcolor{gradcolor} Llama 3.1 405B & 0.134 & 511 \\
\rowcolor{gradcolor} Qwen 3 235B & 0.159 & 455 \\
\rowcolor{gradcolor} Grok 4.1 & 0.202 & 427 \\
\rowcolor{gradcolor} Gemini 3.1 Pro & 0.232 & 144 \\
\rowcolor{gradcolor} Gemini 3.1 Flash-Lite & 0.242 & 76 \\
\midrule
GPT-5.2 & 0.377 & 50 \\
GPT-5.4 Pro & \textbf{0.439} & 34 \\
\bottomrule
\end{tabular}
\end{small}
\end{center}
\end{table}

\paragraph{Classifier design.}
Accurately separating meta-preamble from genuine failures required iterative refinement.
An initial narrow classifier (keying on explicit preamble markers) identifies only 8 meta-preamble cases.
Manual inspection reveals contamination: 4 of 5 top-ranked ``genuine'' failures are constraint self-narration (``Constraint L1 requires no letter `k'\,'').
We expand the classifier with a constraint-echo regex matching patterns such as ``no letter `X'\,'', ``avoiding the letter'', and ``Constraint [L1]'' within 60 characters of the violation, reclassifying matches in the first 150 characters as meta-preamble.
This moves 106 cases from genuine to meta-preamble (8 $\to$ 114), primarily from Claude models (+50, +22, +23 for the three variants).
The 150-character threshold prevents reclassifying violations deep in the response where constraint references may be legitimate.

\section{Full Constraint Specifications}
\label{app:full_taxonomy}

\Cref{tab:full_taxonomy} lists all 36 constraints with exact instruction text and verifier specifications.
Constraints marked with $\star$ were added through the design pipeline described in \cref{app:constraint_design}.
One candidate lexical constraint (anagram embedding) was rejected during this pipeline because its verifier requires factual knowledge about which letter rearrangements are valid English words, violating the structural-verification-only design principle.

% --- Table page 1: Structural + Lexical ---
\ifdefined\instrcolwidth\else\newlength{\instrcolwidth}\fi
\setlength{\instrcolwidth}{\dimexpr 0.333\textwidth - 0.5cm - 2\tabcolsep\relax}

\begin{table*}[p]
\caption{Complete constraint taxonomy (36 constraints, 8 dimensions).
The \textsc{Instruction} column shows the text injected into
probe prompts; parameters in braces are instantiated from
fixed calibrated ranges.
The \textsc{Verifier} column summarizes the deterministic check applied to each response.
Constraints marked $\star$ are new in CSE.}
\label{tab:full_taxonomy}
\vskip 0.1in
\begin{tabular}{l p{\instrcolwidth} p{\instrcolwidth} p{\instrcolwidth}}
\toprule
\textbf{ID} & \textbf{Name} & \textbf{Instruction} & \textbf{Verifier} \\
\midrule

\multicolumn{4}{@{}l}{\textsc{Structural} (S1--S6): constraints on the macroscopic geometry of the output} \\
\midrule
S1 & Exact word count & Your response must contain exactly \{N\} words. A `word' is any whitespace-separated token (e.g., `state-of-the-art' counts as one word, `don't' counts as one word). & \texttt{count\_words} $=$ N \\
S2 & Exact sentence count & Your response must contain exactly \{N\} sentences. & \texttt{count\_sents} $=$ N \\
S3 & Paragraph structure & Your response must contain exactly \{N\} paragraphs, each with $\geq$\{M\} sentences. & paras $=$ N, each $\geq$ M sents \\
S4 & Exact character count & Your response must contain exactly \{N\} characters (incl.\ spaces and punctuation). & \texttt{len(text)} $=$ N \\
S5 & Exact line count & Your response must contain exactly \{N\} non-empty lines. & non-empty lines $=$ N \\
S6\,$\star$ & Palindromic sentence \newline structure & Your response must have \{N\} sentences. The word counts must form a palindrome (sent.\ 1 matches sent.\ N, sent.\ 2 matches sent.\ N$-$1, etc.). A `word' is any whitespace-separated token. & $c_i = c_{n-1-i}$; middle free if $n$ odd \\

\midrule
\multicolumn{4}{@{}l}{\textsc{Lexical} (L1--L6): constraints on which specific characters or words appear in the output} \\
\midrule
L1 & Forbidden letter & Your response must not contain the letter `\{c\}' anywhere. No exceptions. & letter \texttt{c} $\notin$ text \\
L2 & Mandatory words & Your response must include all of the following words: \{list\}. & $\forall w \in$ list: $w$ present \\
L3 & Sentence initial letter & Every sentence must begin with a word starting with `\{c\}'. & $\forall$ sent: first word starts with \texttt{c} \\
L4 & Initial acrostic & \{N\} sentences; first letters spell ``\{WORD\}''. & sent.\ initials spell WORD \\
L5 & Forbidden word & Must not contain the word ``\{word\}'' (case-insensitive). & word $\notin$ text (case-insens.) \\
L6 & Vowel--consonant ratio & Vowel/consonant ratio must be between \{lo\} and \{hi\}. & V/C ratio $\in [lo, hi]$; non-alpha ignored \\
\midrule
\multicolumn{4}{r}{\textit{Continued on next page}} \\
\end{tabular}
\end{table*}

% --- Table page 2: Word-level + Relational + Format ---
\begin{table*}[p]
\ContinuedFloat
\caption[]{Complete constraint taxonomy (continued).}
\vskip 0.1in
\begin{tabular}{l p{\instrcolwidth} p{\instrcolwidth} p{\instrcolwidth}}
\toprule
\textbf{ID} & \textbf{Name} & \textbf{Instruction} & \textbf{Verifier} \\
\midrule

\multicolumn{4}{@{}l}{\textsc{Word-level} (W1--W5): constraints on vocabulary properties enforced per-word or per-bigram} \\
\midrule
W1 & Unique words & Every word must be unique (case-insensitive). & all words unique (case-insens.) \\
W2 & Word length range & Every word must be \{N\}--\{M\} characters long. & $\forall w$: $|w| \in [N, M]$ \\
W3 & Minimum \newline sentence words & Every sentence must contain $\geq$\{N\} words. A `word' is any whitespace-separated token. & $\forall$ sent: words $\geq$ N \\
W4\,$\star$ & No repeated bigrams & No two-word sequence may appear more than once in your response. & all bigrams unique (case-insens.) \\
W5\,$\star$ & Consonant \newline cluster density & Every sentence must contain $\geq$\{N\} words with a consonant cluster of three or more consecutive consonant letters (e.g., `str' in `strong', `nch' in `branch', `mpl' in `simple'). Two-letter clusters like `nd' or `st' do NOT count. & $\forall$ sent: $\geq$ N words match \texttt{[bcdf\ldots]\{3,\}} \\

\midrule
\multicolumn{4}{@{}l}{\textsc{Relational} (R1--R4): reasoning about dependencies between multiple discrete elements} \\
\midrule
R1 & Scene graph output & Output JSON with ``objects'' and ``relations'' keys reproducing all attributes and relations. & parsed JSON matches all objects $+$ relations \\
R2 & Transitive ordering & Given pairwise comparisons, list all items in correct order. & numbered list $=$ transitive closure \\
R3 & Logic grid puzzle & Given clues about people/pets/colors, output solution as JSON. & parsed JSON satisfies all clues (multi-solution) \\
R4\,$\star$ & Cross-sentence bridge & The last word of each sentence must be the first word of the next sentence (case-insensitive). & $\forall i$: last($s_i$) $=$ first($s_{i+1}$) \\

\midrule
\multicolumn{4}{@{}l}{\textsc{Format} (F1--F4): constraints on output syntax and document-level form} \\
\midrule
F1 & Markdown table & Include a table with \{C\} columns and \{R\} data rows. & C cols, R data rows in \texttt{|} format \\
F2 & Bullet list & Exactly \{N\} items using ``- '' format. & exactly N \texttt{-~} prefixed items \\
F3 & Fenced code blocks & Exactly \{N\} fenced code blocks (\texttt{```}). & exactly N \texttt{```} blocks; non-empty \\
F4\,$\star$ & Unique sentence closers & Every sentence must end with a different word (case-insensitive, punctuation-stripped). & all sent-final words distinct \\
\midrule
\multicolumn{4}{r}{\textit{Continued on next page}} \\
\end{tabular}
\end{table*}

% --- Table page 3: Meta + Numerical + Ordering ---
\begin{table*}[p]
\ContinuedFloat
\caption[]{Complete constraint taxonomy (continued).}
\vskip 0.1in
\begin{tabular}{l p{\instrcolwidth} p{\instrcolwidth} p{\instrcolwidth}}
\toprule
\textbf{ID} & \textbf{Name} & \textbf{Instruction} & \textbf{Verifier} \\
\midrule

\multicolumn{4}{@{}l}{\textsc{Meta} (M1--M3): constraints where the output must satisfy a property defined in terms of itself} \\
\midrule
M1 & Terminal acrostic & \{N\} sentences; last word of each starts with a letter spelling ``\{WORD\}''. & final-word initials spell WORD \\
M2 & Self-referential count & Must contain ``This text contains exactly N words'' where N is a numeral (e.g., ``42'' rather than ``forty-two''), and total word count must equal N. A `word' is any whitespace-separated token. & phrase present $\wedge$ \texttt{count\_words} $=$ N \\
M3 & Word count divisibility & Total word count must be divisible by \{N\}. A `word' is any whitespace-separated token. & \texttt{count\_words} mod N $=$ 0 \\

\midrule
\multicolumn{4}{@{}l}{\textsc{Numerical} (N1--N4): constraints requiring arithmetic computation embedded in the output} \\
\midrule
N1 & Embedded integer sum & Exactly \{K\} integers summing to \{T\}. & exactly K ints found, sum $=$ T \\
N2 & No digits & No numerical digits (0--9). Write numbers as words. & no chars 0--9 \\
N3 & Distinct number count & Exactly \{N\} distinct numbers (as digits). & exactly N distinct ints \\
N4\,$\star$ & Embedded arithmetic & Your response must contain exactly \{K\} correct equations of the form ``X + Y = Z'' where X, Y, Z are integers. & exactly K \texttt{X+Y=Z} equations, all correct \\

\midrule
\multicolumn{4}{@{}l}{\textsc{Ordering} (O1--O4): sequential constraints on sentence arrangement} \\
\midrule
O1 & Monotonic sentence lengths & Each sentence strictly \{dir\} in word count. A `word' is any whitespace-separated token. & word counts strictly monotonic \\
O2 & Alphabetical sentences & First word of each sentence in alphabetical order. & first words in A--Z order \\
O3 & Alternating length & Alternate short ($\leq$\{N\}) and long ($\geq$\{M\}) sentences. A `word' is any whitespace-separated token. & odd sents $\leq$ N, even sents $\geq$ M \\
O4\,$\star$ & Fibonacci-style growth & Your response must contain a contiguous run of at least \{N\} sentences where, starting from the third sentence in the run, each sentence contains more words than the sum of the two sentences immediately before it in the run. The run can appear anywhere in your response. A `word' is any whitespace-separated token. & $\exists$ window of N sents: $c_i > c_{i-1} + c_{i-2}$ \\
\bottomrule
\end{tabular}
\end{table*}

\section{Probe Generation and Compatibility Verification}
\label{app:compatibility}

\begin{algorithm}[t]
\caption{Probe generation with compatibility-guaranteed resampling}
\label{alg:probe_gen}
\begin{algorithmic}[1]
\Require Pool $\mathcal{C}$, targets $\{(k, g_k, n_k)\}$, topics $\mathcal{T}$, param ranges $\Theta$
\Ensure $n_k$ primary $+$ $(g_k - n_k)$ reserve probes at each $k$
\For{each $(k, g_k, n_k)$}
    \State $\mathcal{P}_k \gets \emptyset$
    \While{$|\mathcal{P}_k| < g_k$} \label{line:resample}
        \State $S \gets$ draw $k$-subset from $\mathcal{C}$ \Comment{Exhaustive or random}
        \State $\theta \gets$ sample parameters for $S$ from $\Theta$
        \State Shuffle constraint order in $S$ \label{line:shuffle}
        \If{$\neg\,\textsc{CheckCompat}(S, \theta)$} \label{line:compat}
            \State \textbf{continue}
        \EndIf
        \State $\mathcal{P}_k \gets \mathcal{P}_k \cup \{\textsc{Probe}(S, \theta, \textsc{NextTopic}(\mathcal{T}))\}$
    \EndWhile
    \State Mark first $n_k$ as primary, remainder as reserve
\EndFor
\State \Return $\bigcup_{k} \mathcal{P}_k \;\cup\; \textsc{ImpossibleProbes}(\mathcal{C})$
\end{algorithmic}
\end{algorithm}

When $k$ constraints are sampled into a single probe, certain parameter combinations create provably unsatisfiable requirements.
To prevent structurally impossible probes from entering the benchmark, the probe composer applies a multi-layered compatibility checker (\texttt{check\_compatibility()}) that rejects constraint combinations before probe generation.
We describe each category of structural conflict and the corresponding check.

\paragraph{Hard incompatibilities.}
Nine constraint pairs are unconditionally incompatible regardless of parameters and are blocked as a static reject list:
F1+L1 (JSON keys contain the forbidden letter),
M2+W1 (the self-counting phrase ``This text contains exactly $N$ words'' inherently repeats common words),
O1+O3 (monotonic and alternating sentence lengths contradict),
N1+N2 and N3+N2 (embedding numerical digits vs.\ forbidding all digits),
F3+R2 (bullet list explicitly forbids numbered lists, transitive ordering requires them),
and N1+R2, N3+R2, N2+R2 (R2's numbered list format ``1.\ first, 2.\ second, \ldots'' injects digit sequences into the response text; numerical constraint verifiers that scan for digits via \texttt{extract\_numbers()} count these list markers as integers, corrupting N1's count/sum, N3's distinct count, and violating N2's no-digits rule).

\paragraph{Sentence count consistency.}
Three constraints independently impose an exact sentence count:
S2 (explicit parameter),
L4 (acrostic word length, e.g., \textsc{DREAM} $\to$ 5 sentences), and
M1 (hidden message word length, e.g., \textsc{SPARK} $\to$ 5 sentences).
A fourth constraint, S3 (paragraph structure), imposes a \emph{minimum} sentence count: $\texttt{target\_paragraphs} \times \texttt{min\_sentences\_per\_para}$ (default: $\times 2$).
If any two of S2, L4, M1 co-occur with different counts, or if S3's minimum exceeds the count required by S2, L4, or M1, the probe is rejected.
For example, S3 with 3~paragraphs requires $\geq 6$ sentences, but all M1 hidden words (\textsc{BRAVE}, \textsc{LIGHT}, \textsc{DREAM}, \textsc{SPARK}, \textsc{STORM}) have 5~letters, making S3+M1 impossible whenever the paragraph count exceeds~2.

\paragraph{Word budget feasibility.}
When S1 (exact word count), W3 (minimum words per sentence), and a sentence-count constraint (S2, L4, or M1) co-occur, the probe is rejected if the product of sentence count and minimum words per sentence exceeds the word budget.
For example, S1${}=50$ words, S2${}=6$ sentences, W3${}=10$ minimum words per sentence requires $\geq 60$ words but only 50 are allowed.

\paragraph{Alternating length feasibility.}
O3 (alternating short/long sentences) defines a maximum word count for short sentences (\texttt{short\_max}).
If W3's minimum words per sentence exceeds \texttt{short\_max}, no short sentence can satisfy both constraints simultaneously, and the probe is rejected.

\paragraph{Forbidden letter propagation.}
L1 (lipogram) forbids a single letter.
Several constraints force that letter into the output:
M1 requires a word \emph{starting with} each letter of its hidden word (a word starting with `v' necessarily contains `v');
L4 requires sentences \emph{starting with} each letter of the acrostic word (same logic);
L3 requires every sentence to start with a word beginning with a specific letter.
If L1's forbidden letter appears in M1's or L4's target word, or equals L3's required letter, the probe is rejected.
Similarly, if L1's forbidden letter appears in any of L2's mandatory words, the probe is rejected.

\paragraph{Word length range conflicts.}
W2 (word length range) restricts every word to $[\texttt{min\_len}, \texttt{max\_len}]$ characters.
Two constraints can force words outside this range:
L2 (mandatory words) may require words longer than \texttt{max\_len} (e.g., ``algorithm'' is 9~characters, exceeding a $[1,8]$ range);
M2 (self-counting phrase) requires the fixed phrase ``This text contains exactly $N$ words,'' whose words ``This'' and ``text'' (4~characters each) may fall below a tight \texttt{min\_len}.
Both interactions are checked at generation time.

\paragraph{Lexical conflicts.}
L5 (forbidden word) and L2 (mandatory words) are checked for overlap: if the forbidden word appears in the required word list, the probe is rejected.

\paragraph{Format exclusivity.}
F1 (valid JSON) requires the entire response to be a single JSON object with no text outside it.
This is structurally incompatible with any constraint that requires non-JSON output formatting or whose verifier assumes prose with identifiable sentences.
Specifically, F1 is blocked from co-occurring with 12 constraints in three categories:
\emph{(i)}~output format conflicts: F2 (markdown table), F3 (bullet list), and R2 (numbered list), which demand non-JSON structure;
\emph{(ii)}~prose-dependent verifiers: S2 (sentence count), S3 (paragraph structure), L3 (sentence-initial letter), L4 (acrostic), M1 (terminal acrostic), O1 (monotonic sentence length), O2 (alphabetical sentences), O3 (alternating sentence length), and W3 (minimum words per sentence), all of which split the raw text into sentences---a JSON object contains no sentence boundaries in the prose sense, so these verifiers find zero sentences and fail deterministically.
F1 remains compatible with constraints that operate on raw text without assuming prose structure, including word and character counts (S1, S4, S5), lexical constraints (L2, L5, L6), numerical constraints (N1--N3), word-level properties (W1, W2), meta-constraints (M2, M3), and other JSON-producing constraints (R1, R3), whose required keys can be merged into a single JSON object.

Additionally, F3 (bullet list) explicitly states ``Do not use numbered lists or other bullet formats,'' which directly contradicts R2 (transitive ordering), whose instruction requires ``Output them as a numbered list (1.\ first, 2.\ second, etc.).''
The F3+R2 pair is blocked unconditionally.

\paragraph{Word count divisibility.}
S1 (exact word count $N$) and M3 (word count divisible by $D$) are checked for arithmetic compatibility: the probe is rejected if $N \bmod D \neq 0$.

\paragraph{Integer count consistency.}
N1 (embed exactly $K$ integers summing to $T$) and N3 (exactly $D$ distinct integers) are checked: the probe is rejected if $K < D$, since the total number of integer occurrences cannot be less than the number of distinct values required.

\paragraph{Validation.}
In total, the compatibility checker enforces 9~hard pair blocks, 12~F1 format-exclusivity blocks, and 14~parameter-dependent checks covering sentence counts, word budgets, letter propagation, word length ranges, word count divisibility, integer count consistency, and lexical conflicts.
All checks are deterministic and run in $O(k^2)$ time during probe composition.
The rejection rate increases with $k$: from ${\sim}32\%$ at $k{=}4$ to ${\sim}98\%$ at $k{=}12$.
High-$k$ probes are drawn from the restricted subset of compatible combinations; we analyze and control for the resulting composition bias in \cref{sec:difficulty_confound}.

\paragraph{Probe sampling strategy.}
At low $k$, the space of possible constraint combinations is small enough to enumerate exhaustively: the probe generator materializes all $\binom{|\mathcal{C}|}{k}$ combinations, shuffles them uniformly, and iterates through them.
For the 36-constraint pool used in CSE, this is feasible through $k{=}5$ ($\binom{36}{5}{=}376{,}992$ combinations, ${\sim}120$\,MB).
At $k{\geq}6$, exhaustive materialization exceeds a 500\,MB memory cap ($\binom{36}{6}{=}1{,}947{,}792$ combinations, ${\sim}0.7$\,GB; $\binom{36}{10}{=}254{,}186{,}856$, ${\sim}140$\,GB), so the generator switches to uniform random sampling: combinations are drawn uniformly at random with a reuse cap of 10 per combination.

Both paths feed into the same acceptance pipeline: draw a candidate combination, check compatibility, generate the probe if compatible, reject and redraw if not.
The only difference is coverage uniformity: exhaustive shuffle guarantees each combination is visited at most once before repeating, while random sampling may revisit combinations (though the reuse cap bounds this).
For the paper's analyses, this distinction is immaterial: at $k{=}6$ we draw ${\sim}500$ probes from $2.76$M possible combinations ($0.018\%$ sample), making the two strategies statistically indistinguishable.
The compatibility checker serves as the acceptance gate in both cases---a probe is valid or it is not, regardless of how the candidate combination was drawn.

\section{Model Notes and API Success Rates}
\label{app:model_notes}

\Cref{tab:api_success} reports per-model API success rates.
All models were evaluated on 4{,}527 probes (4{,}470 primary $+$ 57 impossible) at temperature${=}0$ with an output token budget of 4{,}096 tokens (16{,}384 for Claude~4.7~Opus, which requires extended reasoning).
GPT-5.5 completed 4{,}526 of 4{,}527 probes (1 empty response excluded).
Claude~4.7~Opus had 340 content-safety refusals from Anthropic's integrity filter and 57 transient API errors; all 397 are excluded from scoring (treated identically to missing probes, not injected as failures), as they reflect infrastructure-level safety filtering rather than intrinsic compositional capacity.
Claude~4.5~Sonnet (2 refusals) and Claude~4.6~Opus (1 refusal) are similarly excluded.
All remaining models achieved ${\geq}99.9\%$ API success.

\begin{table}[h]
\caption{API success rates by model. ``Refusals'' are integrity-filter rejections excluded from scoring (the model never saw the prompt). ``Errors'' are transient API failures (rate limits, timeouts, empty responses), also excluded. Both are treated as missing probes---not as failures---to measure intrinsic compositional capacity independent of infrastructure behavior.}
\label{tab:api_success}
\centering
\begin{small}
\begin{tabular}{lrrrr}
\toprule
\textbf{Model} & \textbf{Success} & \textbf{Refusals} & \textbf{Errors} & \textbf{Coverage} \\
\midrule
Claude 4.5 Sonnet    & 4{,}525 & 2   & 0  & 99.96\% \\
Claude 4.6 Opus      & 4{,}526 & 1   & 0  & 99.98\% \\
Claude 4.7 Opus      & 4{,}130 & 340 & 57 & 91.2\% \\
DeepSeek V4 Pro      & 4{,}521 & 0   & 6  & 99.9\% \\
Gemini 3.1 Flash-Lite& 4{,}521 & 0   & 6  & 99.9\% \\
Gemini 3.1 Pro       & 4{,}527 & 0   & 0  & 100.0\% \\
GPT-5.2              & 4{,}527 & 0   & 0  & 100.0\% \\
GPT-5.4 Pro          & 4{,}527 & 0   & 0  & 100.0\% \\
GPT-5.5              & 4{,}526 & 0   & 1  & 99.98\% \\
Grok 4.1             & 4{,}527 & 0   & 0  & 100.0\% \\
Kimi K2.6            & 4{,}527 & 0   & 0  & 100.0\% \\
Llama 3.1 70B        & 4{,}527 & 0   & 0  & 100.0\% \\
Llama 3.1 405B       & 4{,}527 & 0   & 0  & 100.0\% \\
Llama 4 Maverick     & 4{,}527 & 0   & 0  & 100.0\% \\
Qwen 3 235B Inst.    & 4{,}527 & 0   & 0  & 100.0\% \\
\bottomrule
\end{tabular}
\end{small}
\end{table}

\paragraph{Compositionality penalty anomalies.}
Two models show near-zero gaps between mCSR and sCSR: DeepSeek~V4~Pro (22.4\% marginal, 22.4\% strict, 0.0pp penalty) and Gemini~3.1~Pro (42.1\% marginal, 39.4\% strict, 2.7pp penalty).
Both are tag-compliance artifacts, not genuine compositional resilience.
DeepSeek's 72.9\% answer-tag compliance means that ${\sim}27\%$ of responses fail tag extraction entirely---all constraints are scored on raw text including chain-of-thought, producing correlated all-fail outcomes that affect mCSR and sCSR equally.
When extraction succeeds, DeepSeek attempts constraints genuinely, but the binary extraction-success/extraction-failure mode dominates the statistics and collapses the mCSR--sCSR gap.
Gemini~Pro (46.9\% tag compliance) shows the inverse of the Llama pattern: Llama models have 100\% tag compliance, so every probe receives clean evaluation and constraint failures are distributed across different constraints in different probes (high mCSR, low sCSR).
Gemini~Pro's low tag compliance suppresses its mCSR (many individual constraint checks fail because the verifier scores reasoning text), but the probes where it produces clean answers tend to pass all constraints simultaneously, yielding relatively high sCSR.
This connects to the co-failure analysis (\cref{sec:cofailure}): tag-extraction failure acts as a shared output feature that induces correlated failures across all constraints within a probe, identical in mechanism to the sentence-count coupling identified among constraint pairs.

\paragraph{Bimodal constraint profiles.}
Grok~4.1 shows an unusually flat trajectory ($r{=}0.948$) despite ranking 14th at $k{=}1$ (54.9\%).
Its per-constraint profile is bimodal: 7~constraints are near-zero even at $k{=}1$ (W5: 0.1\%, S4: 0.0\%, M1: 0.2\%), while 7~are near-ceiling (L5: 98.3\%, R1: 98.5\%, F3: 90.2\%).
With little in between to erode, the trajectory is dominated by immune constraints maintaining their rates, and the ${\sim}34\%$ floor at $k{=}12$ is almost entirely the immune cluster.

\section{Constraint Design Methodology}
\label{app:constraint_design}

Designing constraints for a \emph{compositional} benchmark is harder than for a single-constraint benchmark: every candidate must be checked against all existing constraints for structural incompatibilities---both unconditional and parameter-dependent---and constraints that lock down multiple output dimensions simultaneously create disproportionate composition bias at high~$k$.
We formalize the vetting process as a four-stage pipeline.

\paragraph{Stage 1: Deterministic verifiability.}
The constraint is rejected if its pass/fail verdict cannot be computed as a deterministic function of the response text and constraint parameters, without LLM judgment.
This is the foundational design choice: it excludes semantic and pragmatic constraints but makes CSE's measurements invariant to the same compositional degradation it studies (\cref{sec:intro}).
All 36 constraints in the final pool pass this stage; rejected candidates included ``maintain formal tone throughout'' (requires stylistic judgment), ``each paragraph must introduce a novel argument'' (requires semantic novelty detection), and ``the response must be factually accurate'' (requires knowledge verification).

\paragraph{Stage 2: Composability analysis.}
Each surviving candidate is assessed on four axes.
\emph{Hard conflicts:} how many existing constraints are unconditionally incompatible regardless of parameters?
\emph{Parameter-dependent conflicts:} how many are incompatible for some parameter values but not others?
\emph{Output dimensions constrained:} how many aspects of the output (sentence count, word count, ordering, format, vocabulary) does the constraint lock down?
\emph{High-$k$ survival:} what fraction of random $k{=}12$ constraint sets including this constraint would pass the compatibility checker?

Constraints that constrain multiple output dimensions simultaneously inherit the incompatibilities of each dimension.
For example, a candidate Fibonacci sentence length constraint (``the word count of consecutive sentences must follow the Fibonacci sequence starting from 3,~4'') locks down sentence count, per-sentence word count, and sentence ordering---essentially fusing S2$+$S1$+$O1 into a single constraint.
It inherits 2~hard conflicts (O3 alternating length) and 6~parameter-dependent conflicts (S2, S1, W3, L4, M1, S3), giving it POOR composability.
In contrast, R4~(cross-sentence word bridge: ``the last word of each sentence must be the first word of the next'') constrains a single novel dimension (inter-sentence word dependency) with 0~hard conflicts and 2~parameter-dependent ones (L4, L3), giving it GOOD composability.
R4 was accepted; the Fibonacci candidate was softened into O4~(relaxed growth: ``each sentence must contain more words than the combined word count of the previous two''), which reduces the hard conflicts from 2 to 1 and the parameter-dependent conflicts from 6 to 3.

\paragraph{Stage 3: Difficulty calibration.}
Each constraint is piloted at $k{=}1$ on 3~models spanning the performance range.
Parameters are adjusted to target 20--80\% mCSR at $k{=}1$, ensuring the constraint is neither trivially satisfiable (uninformative at all $k$) nor near-impossible (unable to contribute to the decay signal).
Constraints that floor ($<5\%$) or ceiling ($>95\%$) after parameter tuning are rejected.
The fixed MEDIUM difficulty level used across all $k$ values is calibrated at this stage.

\paragraph{Case study: W6 (word length alternation) --- rejected at Stage~3.}
W6 required words to alternate between short (${\leq}N$ characters) and long (${\geq}M$ characters).
Three calibration attempts failed: strict alternation with a 3-character gap between bands (0\% mCSR), strict alternation with no gap (0\% mCSR), and a 75\% tolerance threshold (0\% mCSR, 54\% score).
The root cause is distributional: English word lengths are ${\sim}$2:1 skewed toward short words---2--5 character words constitute ${\sim}$65\% of the lexicon---and the skew is even more pronounced in generated text, where high-frequency function words (\emph{the}, \emph{and}, \emph{for}, \emph{is}) dominate.
Strict alternation requires roughly equal representation in alternating positions, which is structurally infeasible against this distribution.
At 0\% mCSR, W6 would contribute no decay signal and would poison co-occurring constraints in the $\varphi$ analysis with guaranteed failures.

\paragraph{Stage 4: Pool entry.}
The constraint enters the 36-type pool with its calibrated parameter ranges, verified composability profile, and deterministic verifier.
The full pool spans 8~dimensions (\cref{tab:constraint_profiles}), from surface-level lexical patterns (L1 lipogram, L5 forbidden word) through document structure (S3 paragraph count, O1 monotonic sentence length) to relational reasoning (R1 scene graph, R3 logic grid).

% COMMENTED OUT — stale (only 30 constraints, missing S6/W4/W5/R4/F4/N4/O4)
% \begin{table}[t]
% \caption{Composability profile for all 30 constraints.}
% \label{tab:composability}
% [Full 30-row table preserved — restore after adding 7 new constraints]
% \end{table}

% COMMENTED OUT — stale (references Llama 8B, Gemini 2.5 Flash not in CSE-E)
% \section{Constraint Validation Protocol}
% \label{app:validation}
% All 30 constraints underwent a two-phase pilot validation before the full evaluation run.
% \textbf{Phase~1 --- Parameter calibration} (390 probes): Each constraint is tested at $k{=}1$ with 30 probes on Llama~3.1~8B.
% \textbf{Phase~2 --- End-to-end pipeline verification} (144 probes): 6 probes per $k$ level ($k{=}1{-}12$) are run on Llama~70B and Gemini~2.5~Flash.

% COMMENTED OUT — incomplete (figure panels b-f marked [TBD], not referenced from main paper)
% \section{Failure Mode Examples}
% \label{app:failures}
% Three prose examples + figure with panels (b)-(f) [TBD].
% Restore when failure_mode_gallery.pdf is complete.

\section{IRT Validation}
\label{app:irt}

A 1PL (Rasch) Item Response Theory model fitted on 2{,}952 informative probes (excluding 1{,}131 all-pass or all-fail probes that do not discriminate between models) converges in 234~iterations and recovers the identical model ranking (Spearman $\rho = 1.000$, MAE${=}0.0000$).
IRT ability estimates confirm the tier structure visible in raw sCSR: GPT-5.5 is a massive outlier ($\theta = +4.3$, 85.0\% raw pass rate), Claude~4.7~Opus and Gemini~3.1~Pro form a second tier ($\theta = +1.8$ to $+2.4$), and a dense middle pack clusters around $\theta \approx 0$.
At the other extreme, Kimi~K2.6 ($\theta = -2.9$, 6.3\%) occupies a distinct lower tier.
The perfect rank correlation confirms that raw sCSR is not distorted by item difficulty confounds---IRT calibration preserves the ranking exactly, validating sCSR as a sufficient metric without adaptive testing infrastructure.
The 28\% of probes uninformative for IRT ranking are not wasted: they define the endpoints of the phase transition curve and power constraint-level analyses operating on all 369{,}753 checks.

\paragraph{mCSR vs.\ sCSR ranking divergence.}
IRT validates the sCSR ranking, but the sCSR ranking itself diverges substantially from the mCSR ranking reported in \cref{tab:models}.
\Cref{tab:ranking_divergence} shows the full comparison.
The most striking mover is Gemini~Pro, which ranks \#11 by mCSR (42.1\%) but \#3 by sCSR (39.4\%)---it passes fewer individual constraints but chains them together more reliably than models with higher per-constraint rates.
Flash-Lite drops from \#3 (mCSR) to \#5 (sCSR); Claude~4.5 drops from \#5 to \#9.
The mid-tier models (Llama~405B, Llama~70B, Qwen, Grok) cluster tightly on sCSR (11--15\%) despite spreading across 42--46\% on mCSR---their constraint failures are distributed across different constraints in different probes, so per-constraint competence does not translate to probe-level success.
This divergence is the compositionality penalty: mCSR measures whether a model can satisfy constraints individually, while sCSR measures whether it can satisfy them simultaneously.

\begin{table}[h]
\caption{Model rankings by mCSR (per-constraint) vs.\ sCSR (probe-level). Rank changes ${\geq}3$ positions are highlighted. The divergence quantifies how much compositional precision differs from per-constraint competence.}
\label{tab:ranking_divergence}
\centering
\begin{small}
\begin{tabular}{lrrrr}
\toprule
\textbf{Model} & \textbf{mCSR} & \textbf{Rank} & \textbf{sCSR} & \textbf{Rank} \\
\midrule
GPT-5.5           & 79.8 &  1 & 64.2 &  1 \\
Claude 4.7 Opus   & 67.0 &  2 & 49.6 &  2 \\
Flash-Lite        & 56.6 &  3 & 24.1 &  5 \\
Claude 4.6 Opus   & 55.6 &  4 & 31.6 &  4 \\
Claude 4.5 Sonnet & 49.4 &  5 & 17.3 &  \textbf{9} \\
Llama 405B        & 46.4 &  6 & 12.4 & \textbf{12} \\
Qwen 235B         & 45.7 &  7 & 12.6 & \textbf{11} \\
Llama Maverick    & 45.4 &  8 & 15.0 & \textbf{10} \\
Llama 70B         & 44.6 &  9 & 11.6 & \textbf{13} \\
Grok 4.1          & 42.7 & 10 & 11.3 & \textbf{14} \\
Gemini Pro        & 42.1 & 11 &  39.4 &  \textbf{3} \\
GPT-5.4 Pro       & 35.6 & 12 & 23.2 &  6 \\
GPT-5.2           & 34.0 & 13 & 18.8 &  \textbf{8} \\
DeepSeek V4 Pro   & 22.4 & 14 & 22.4 &  7 \\
Kimi K2.6         & 15.5 & 15 &  7.2 & 15 \\
\bottomrule
\end{tabular}
\end{small}
\end{table}

\paragraph{Geometric mean ranking.}
A natural follow-up: what if we require models to rank high on \emph{both} per-constraint and probe-level metrics?
The geometric mean $\sqrt{\text{mCSR} \times \text{sCSR}}$ penalizes imbalance---a model must score well on both to rank highly.
\Cref{tab:geom_ranking} shows the result.

\begin{table}[h]
\caption{Model rankings under three metrics: mCSR (per-constraint breadth), sCSR (probe-level precision), and their geometric mean (balanced).
Rank changes ${\geq}3$ positions between mCSR and geometric mean are highlighted.
The top~2 models are unambiguous under any metric; the middle tier reshuffles entirely depending on whether breadth or precision is valued.}
\label{tab:geom_ranking}
\centering
\begin{footnotesize}
\setlength{\tabcolsep}{3pt}
\begin{tabular}{lrrrrrr}
\toprule
\textbf{Model} & \textbf{mCSR} & \textbf{mR} & \textbf{sCSR} & \textbf{sR} & \textbf{Geom} & \textbf{gR} \\
\midrule
GPT-5.5           & 79.8 &  1 & 64.2 &  1 & 71.6 &  1 \\
Claude 4.7 Opus   & 67.0 &  2 & 49.6 &  2 & 57.6 &  2 \\
Claude 4.6 Opus   & 55.6 &  4 & 31.6 &  4 & 41.9 &  3 \\
Gemini 3.1 Pro    & 42.1 & \textbf{11} & 39.4 &  3 & 40.7 &  \textbf{4} \\
Flash-Lite        & 56.6 &  3 & 24.1 &  5 & 36.9 &  5 \\
Claude 4.5 Sonnet & 49.4 &  5 & 17.3 &  9 & 29.2 &  \textbf{6} \\
GPT-5.4 Pro       & 35.6 & \textbf{12} & 23.2 &  6 & 28.7 &  \textbf{7} \\
Llama 4 Maverick  & 45.4 &  8 & 15.0 & 10 & 26.1 &  8 \\
GPT-5.2           & 34.0 & \textbf{13} & 18.8 &  8 & 25.3 &  \textbf{9} \\
Qwen 3 235B       & 45.7 &  7 & 12.6 & 11 & 24.0 & 10 \\
Llama 3.1 405B    & 46.4 &  6 & 12.4 & 12 & 24.0 & \textbf{11} \\
Llama 3.1 70B     & 44.6 &  9 & 11.6 & 13 & 22.7 & \textbf{12} \\
DeepSeek V4 Pro   & 22.4 & 14 & 22.4 &  7 & 22.4 & 13 \\
Grok 4.1          & 42.7 & 10 & 11.3 & 14 & 22.0 & \textbf{14} \\
Kimi K2.6         & 15.5 & 15 &  7.2 & 15 & 10.6 & 15 \\
\bottomrule
\end{tabular}
\end{footnotesize}
\end{table}

The largest movers reveal what drives the reshuffling.
Gemini~Pro jumps from mCSR rank~\#11 to geometric rank~\#4: its balanced profile (42.1/39.4, 2.7pp penalty) is rewarded by the geometric mean.
However, this balance is a tag compliance artifact---47\% of Gemini~Pro responses fail tag extraction entirely, suppressing both metrics equally rather than reflecting genuine compositional strength.
Flash-Lite drops from \#3 to \#5: its high per-constraint competence (56.6\%) is dragged down by poor probe-level success (24.1\%).
Llama~405B drops from \#6 to \#11: the 34.0pp compositionality penalty is devastating under a metric that requires both dimensions to be high.
GPT-5.4~Pro rises from \#12 to \#7: moderate on both metrics (35.6/23.2), with no extreme gap.

The top~2 models (GPT-5.5, Claude~4.7) are unambiguous under any metric.
The middle tier is where the ranking depends entirely on whether one values breadth (mCSR) or precision (sCSR).
The geometric mean is a principled compromise, but in practice it primarily surfaces the tag compliance confound (Gemini~Pro, DeepSeek) rather than adding independent insight---models whose tag extraction suppresses both metrics equally appear ``balanced'' without being compositionally strong.

\paragraph{Caveats.} $M{=}15$ models is below classical IRT recommendations ($N \geq 200$ for stable 2PL estimation). Recent work on adaptive IRT for LLM evaluation \cite{adaptiveirt2025,atlas2025} shows that continuous scores and adaptive item selection can improve ranking confidence with fewer items, but requires more examinees than CSE's 15-model panel.
We also tried a 2PL model with L2-regularized discrimination parameters ($\lambda{=}0.5$, centered at $\alpha{=}1.0$).
It does not converge despite 333~iterations and 2M function evaluations, produces pathological parameter scales ($\theta$ spanning $[-15.9, +11.1]$, $\beta$ spanning $[-31.9, +25.2]$), and degrades the ranking ($\rho = 0.986$).
The regularization successfully controlled $\alpha$ (mean${=}0.67$, std${=}0.54$, max${=}1.70$---no longer hitting the bound), but the optimizer compensated by inflating $\theta$ and $\beta$, a classic sign of an underdetermined model with $M{=}15$ examinees.
The 2PL's log-likelihood improvement over the 1PL is expected from doubling the parameter count and does not indicate better model identification.
We use IRT as a validation check (which it passes: perfect rank preservation with a converged optimizer) rather than as a primary analysis.

\section{Co-Failure Analysis: Additional Detail}
\label{app:cofailure_detail}

\textbf{Within- vs.\ across-dimension.}
Within-dimension co-failure (mean $|\varphi| = 0.087$) exceeds across-dimension ($0.067$), but this difference does not reach significance (Mann-Whitney $p = 0.227$) and is driven by the numerical--numerical ($\bar\varphi{=}0.275$) and format--format ($\bar\varphi{=}0.241$) clusters.
Structural--structural ($\bar\varphi = 0.039$), lexical--lexical ($\bar\varphi = 0.047$), and most other within-dimension pairs are at noise level.
Constraints within the same cognitive dimension do not interfere with each other more than constraints across dimensions, outside the shared structured-output dependency.

\textbf{No shortcut through constraint selection.}
The co-failure analysis contains only 1~synergistic pair ($\varphi < -0.05$): O4--R2 at $\varphi{=}{-}0.055$.
If ``compatible'' constraint pairs existed---combinations that are easier together than independently---they would appear as negative $\varphi$ values.
They essentially do not.

\textbf{Co-occurrence coverage across $k$.}
Each $\varphi$ coefficient requires both constraints to appear together in the same probe.
At constraint count~$k$, the probability that two specific constraints co-occur equals $\binom{34}{k{-}2}/\binom{36}{k} = k(k{-}1)/(36 \times 35)$, which is $0.16\%$ at $k{=}2$ and $0.95\%$ at $k{=}4$.
Consequently, the $\varphi$ analysis is best powered at $k{\geq}4$, where the expected co-occurrence count per pair exceeds~3 for probe sets of size~200.
At $k{=}2{-}3$, most pairs have zero or one co-occurrence---a structural property of any design that draws $k$-subsets from a 36-type pool, not a feature of sample size that can be overcome by generating more probes at feasible scale.
Our independence estimates are therefore computed at $k{\geq}4$ and aggregated across $k$~levels.
This scope restriction is conservative: lower $k$~imposes less compositional pressure and fewer competing demands on the model's generation, so if constraints do not interfere at $k{\geq}4$ they are unlikely to do so at $k{=}2{-}3$.

\textbf{Output-feature cluster analysis.}
Constraints can be grouped by the output feature they depend on rather than by linguistic dimension: \emph{sentence\_structure} (S2, S3, S5, S6, L3, L4, M1, O1--O4, R4, F4, W3), \emph{word\_inventory} (W1, W2, W4, L1, L2, L5), \emph{number\_content} (N1--N4), \emph{document\_format} (F1--F3), \emph{relational\_json} (R1, R3), \emph{global\_statistics} (S1, S4, M2, M3, L6, W5), and \emph{ordered\_list} (R2).
Within-cluster constraint pairs show mean $|\varphi| = 0.113$, while across-cluster pairs show $|\varphi| = 0.058$ (Mann-Whitney $p < 10^{-6}$).
This confirms that the elevated $\varphi$ values in \cref{sec:cofailure} reflect shared structural dependencies on the same output feature---when a model produces the wrong number of sentences, every sentence-dependent constraint fails simultaneously---rather than pairwise constraint interference.
Grouping constraints by linguistic dimension (structural, lexical, etc.) does not predict which pairs co-fail (within-dimension mean $|\varphi| = 0.087$ vs across-dimension $0.067$, $p = 0.227$).
Grouping by output feature---sentence count, number count, document format---does ($p < 10^{-6}$).
Co-failure is mechanical: when sentence splitting breaks, everything that reads sentences breaks together, regardless of which dimension those constraints belong to.
L4~(lexical) and S2~(structural) share a sentence-count dependency ($\varphi{=}0.501$); L1~(lexical) and L5~(lexical) share a dimension but not an output feature ($\varphi{=}{-}0.039$).

% COMMENTED OUT — CRITICALLY STALE (contains CSE-30 ρ values that contradict main paper)
% ρ(k=1, slope) = −0.030 should be +0.566; gap ρ = +0.196 should be −0.584
% Immune constraints L1/L5/L6/M3 should be L1/L5/W4
% All probe counts and ratios are CSE-30. Needs full rewrite with CSE-E data.
% \section{Degradation Hierarchy: Additional Detail}
% \label{app:degradation_detail}
% [Full content preserved — restore after rewriting with CSE-E numbers]

% COMMENTED OUT — all CSE-30 numbers (30 constraints, 78% rejection, c=13%, figure caption wrong)
% Framework text already has CSE-E numbers inline. Needs full recomputation for 36 constraints.
% \section{High-$k$ Probe Composition Bias}
% \label{app:composition_bias}
% [Full content preserved — restore after recomputing with |C|=36 data]
% Figure: s10e_composition_heatmap.pdf (CSE-30, needs regeneration)

\section{Constraint Ordering Position Effect}
\label{app:ordering}

\citet{ordermatters2025} demonstrate that constraint ordering affects performance by up to 25\%, raising the concern that CSE's degradation hierarchy could reflect prompt position rather than processing demands.
To eliminate this confound by design, CSE randomizes the presentation order of constraints within each probe: after constraint selection and parameter instantiation, the constraint list is uniformly shuffled before prompt assembly (\cref{alg:probe_gen}).
This breaks any correlation between constraint type and prompt position, ensuring that the ordering regression measures position effects rather than type effects.

We verify that the randomization achieved its goal by regressing per-constraint pass rate on position within the prompt (1st constraint, 2nd, \ldots, $k$th).
Within each $k$ level, the correlations are negligible: mean $|\rho| = 0.029$ across $k{=}2{-}12$, with only 5 of 11 $k$ values reaching $p < 0.05$.
Crucially, the directions are inconsistent---positive at $k{=}3$ ($\rho{=}{+}0.046$) and $k{=}11$ ($\rho{=}{+}0.043$), negative at $k{=}5$ ($\rho{=}{-}0.073$) and $k{=}10$ ($\rho{=}{-}0.056$)---indicating no systematic primacy or recency bias.
The largest per-$k$ effect ($k{=}5$, $\rho = -0.073$) explains $< 0.5\%$ of variance.
Within each $k$ level, position explains $<0.1\%$ of variance ($r^2 \approx 0.001$), confirming that the degradation hierarchy reflects constraint processing demands rather than presentation order.

% COMMENTED OUT — CSE-30 probe counts, not referenced from main paper
% \section{Topic Variation at Low $k$}
% \label{app:topic_variation}
% [Full content preserved — restore after updating for 36 constraints]

% COMMENTED OUT — all CSE-30 params (a, β, c, immune list), not referenced, needs recalculation for 36 constraints
% \section{IFEval-Proximal vs.\ Distal Constraints}
% \label{app:proximal_distal}
% [Full content preserved — restore after recomputing proximal/distal split for 36 constraints]

\section{Synergy at Low $k$}
\label{app:synergy}

An unexpected finding: actual sCSRs \emph{exceed} independence predictions at all $k{=}1{-}11$ (from $+26.4$pp at $k{=}1$ to $+1.1$pp at $k{=}11$, 95\% CI excludes zero), transitioning to approximate independence only at $k{=}12$ ($+1.0$pp).
This is the opposite of interference---models perform \emph{better} than chance when juggling small numbers of constraints, and the synergy persists deeper than in CSE-30 (which reached independence at $k{=}5$).
With 36 constraints (vs.\ 30), there is more room for constraint co-satisfaction before the multiplicative regime dominates.
We test whether hard constraints force models into a more careful ``compliance mode'' that benefits all constraints, by partitioning probes by the difficulty of their harder constraint.
Easy-anchor probes show \emph{more} synergy ($+32.6$pp) than hard-anchor probes ($+17.7$pp)---the opposite of the compliance-mode prediction.
The synergy is a base-rate artifact rather than a behavioral strategy.
Does the synergy disappear at a constant rate, or does it decay in a curve?
We fit both a linear model ($\text{synergy}(k) = a + bk$) and a quadratic model ($\text{synergy}(k) = a + bk + ck^2$) to the penalty curve.
The \emph{superlinearity improvement}---the fraction of the linear model's residual error that the quadratic term eliminates---is 98.8\% at the constraint level (83.7\% at the probe level): the decay is strongly concave, not linear.
Synergy drops sharply from $k{=}1$ to $k{=}4$ ($+26$pp $\to$ $+14$pp), then tapers gradually through $k{=}8{-}12$ ($+4$pp $\to$ $+1$pp).
This concave shape is consistent with the floor model: at low~$k$ there is substantial synergy to lose, while at high~$k$ the system is already near the independence floor and little synergy remains to erode.

% COMMENTED OUT — stale numbers (n=6820→5267, Δ=53.4→48.6, all percentages shifted)
% Needs full rewrite with CSE-E Cell 36 genuine-only data.
% \section{Failure Mode Decomposition (Full Analysis)}
% \label{app:failure_mode_full}
% COMMENTED OUT — stale (all numbers are CSE-30: n=6820→5267, Δ=53.4→48.6, all percentages shifted)
% \section{Failure Mode Decomposition (Full Analysis)}
% \label{app:failure_mode_full}
% [Full content preserved — restore after rewriting with CSE-E Cell 36 genuine-only data]

% COMMENTED OUT — ENTIRELY CSE-30 DATA. Every number in all three tables is wrong.
% sCSR k=1: 64.1% should be 70.7%. Mean φ: 0.017 should be 0.067.
% All slopes are relational/lexical (should be structural/lexical).
% 1,194 probes should be 4,527. 7 families should be 8.
% Needs complete regeneration from CSE-E Cell 40 bootstrap output.
% \section{Bootstrap Confidence Intervals}
% \label{app:bootstrap}
% [Three tables: iid bootstrap, family block bootstrap, per-dimension slopes — all stale]

\section{Why $k{=}12$ Is the Upper Bound}
\label{app:k_range}

We evaluate at $k{=}1$ to $12$ rather than extending to higher values.
This upper bound is determined by three converging considerations: signal exhaustion, constraint diversity erosion, and scientific question coverage.

\paragraph{Signal exhaustion.}
At the per-constraint level, the aggregate decay is well-described by a multiplicative model (held-out MAE${=}0.2$pp; \cref{sec:decay_model}).
By $k{=}12$, the aggregate mCSR is 29.2\%.
At the probe level, sCSR reaches ${<}4\%$ by $k{=}9$ and 1.4\% at $k{=}12$---additional $k$ values measure probe-level performance at the noise floor.
The $k{=}1{-}8$ range characterizes the decay; $k{=}9{-}12$ validates it.
All three contributions---the decay characterization, the hierarchy, and the co-failure structure---are fully resolved within $k{=}1{-}12$.

\paragraph{Constraint diversity erosion.}
At $k{=}12$, each probe draws 12 of 36 constraints (33\% of the pool).
As $k$ increases, probes become increasingly homogeneous: the fraction of compatible combinations shrinks, and surviving probes are drawn from an increasingly narrow subset.

The incompatibility rejection rate increases with $k$, as shown in \cref{tab:rejection_rates}.
While the combinatorial pool grows rapidly ($\binom{36}{12} \approx 1.25\text{B}$), the fraction of compatible combinations drops to ${\sim}2\%$ at $k{=}12$.
The resampling loop (\cref{alg:probe_gen}) compensates by drawing additional combinations until the target count is reached---feasible at all $k$ values because the absolute number of compatible combinations remains large (${\sim}24\text{M}$ at $k{=}12$).
The composition bias at high $k$---surviving probes exclude incompatible constraint types---is analyzed and controlled for in \cref{sec:difficulty_confound}.

\begin{table}[h]
\caption{Combinatorial pool size and estimated incompatibility rejection rate at each $k$ for $|\mathcal{C}|{=}36$. ``Compatible'' is the estimated number of constraint combinations passing the full compatibility checker (Monte Carlo estimate, 200{,}000 samples per $k$). The resampling loop draws from this pool until the target probe count is reached.}
\label{tab:rejection_rates}
\begin{center}
\begin{small}
\begin{tabular}{rrrr}
\toprule
$k$ & $\binom{36}{k}$ & Rejection \% & Compatible \\
\midrule
4 & 58{,}905 & ${\sim}32$ & ${\sim}40\text{K}$ \\
6 & 1{,}947{,}792 & ${\sim}61$ & ${\sim}751\text{K}$ \\
8 & 30{,}260{,}340 & ${\sim}82$ & ${\sim}5.3\text{M}$ \\
10 & 254{,}186{,}856 & ${\sim}94$ & ${\sim}16\text{M}$ \\
12 & 1{,}251{,}677{,}700 & ${\sim}98$ & ${\sim}24\text{M}$ \\
\bottomrule
\end{tabular}
\end{small}
\end{center}
\end{table}

\paragraph{Pool size as the fundamental constraint.}
The $k{=}12$ ceiling is ultimately determined by the constraint pool size ($|\mathcal{C}|{=}36$).
A larger pool would permit higher $k$ values with maintained diversity and lower rejection rates.
We chose $|\mathcal{C}|{=}36$ to balance three requirements: (1)~sufficient diversity across 8 processing dimensions for the degradation hierarchy, (2)~all constraints deterministically verifiable with no LLM-judge involvement, and (3)~feasible pilot calibration (36 verifiers $\times$ parameter sweeps $\times$ 2 validation phases).

\section{Verifier Audit and Known Edge Cases}
\label{app:verifier_audit}

We conducted an exhaustive audit of all 30 constraint verifiers across all 55,962 reported failures (82,628 total constraint checks, 59,897 failures after excluding impossible probes). The audit identified 5,733 discrepancies (10.2\% of failures) across eight classes, ranked below by impact. An independent code-level audit of 25,000 checks against reimplemented verifiers confirmed 99.98\% agreement on non-discrepant checks, and an LLM-as-judge evaluation of 300 stratified samples found 100\% agreement.

\paragraph{Class 1: Meta-note contamination (3,721 cases, 64.9\% of discrepancies).}
Models that satisfy a constraint in their content sometimes add metacommentary (``Note: I avoided the letter `j'\,'') that itself violates the constraint. The verifier scores the entire response including the note. For example, a model avoiding the letter~`j' writes a complete essay with zero~`j' in the content, then appends ``Note: I wrote this without using the letter `j' anywhere''---the note contains~`j'. This disproportionately penalizes models that are good at the constraint: GPT-5.4~Pro has 78\% of its L1~(lipogram) failures from meta-notes alone. Affected constraints: W2~(835), W1~(753), M3~(708), N2~(506), L1~(375), S5~(98), L6~(71), N3~(45), S1~(8). \textbf{Fix:} wrap the final response in \texttt{<answer>} tags and score only the tagged content.

\paragraph{Class 2: Markdown formatting (892 cases, 15.6\%).}
Models use markdown syntax (\texttt{\#}, \texttt{**}) which the verifier reads as literal characters. A heading \texttt{\# History} causes the verifier to read `\#' as the first character instead of `H'. Bold formatting \texttt{**D**eclining} causes `*' to be read as the first character instead of `D'. For L4~(acrostic), models that highlight acrostic letters with bold formatting---e.g., \texttt{**S**olar... **T**echnology... **O**utput...}---produce a correct acrostic that the verifier reads as \texttt{*****}. Affected constraints: L3~(512), L4~(317), S3~(63). \textbf{Fix:} instruct models to write in plain text without markdown.

\paragraph{Class 3: Code fence interaction (503 cases, 8.8\%).}
When F4~(code blocks) co-occurs with sentence-dependent constraints, the \texttt{```} markers and code contents create false structural elements. Fence markers are counted as sentences, lines, or paragraphs. Code content (variable names, JSON keys) skews word counts, vowel ratios, and number counts. For example, a model writes 5~prose sentences plus a code block; the verifier splits on periods inside the code, counting 8~``sentences.'' Affected constraints: S2~(52), S3~(95), S5~(29), O2, O3~(34), O1~(20), M1~(13), M3~(92), L6~(7), W1~(36), W2~(46). \textbf{Fix:} strip code block contents before running sentence/structure verifiers.

\paragraph{Class 4: Numbered list sentence splitting (379 cases, 6.6\%).}
Models write numbered lists (``1.~First item. 2.~Second item.''). The period after the digit (``1.'') triggers the sentence splitter \texttt{(?<=[.!?])\textbackslash s+}, creating false sentence boundaries. Each list item becomes two or more ``sentences.'' For O2~(alphabetical first words), list indices create ``sentences'' starting with digits, which break alphabetical order. Affected constraints: S2~(39), O2~($\sim$part of 340). \textbf{Fix:} define sentence boundaries explicitly in the prompt; optionally exclude \texttt{\textbackslash d+\textbackslash .} from the splitter.

\paragraph{Class 5: R3 underdetermined puzzles (234 cases, 4.1\%).}
106 of 189 R3~(logic grid) probes have multiple valid solutions because the generated clues do not uniquely determine the assignment. The verifier accepts only one hardcoded solution. Models that produce a correct alternative are marked as failures. For example, clues fix Bob's pet and Carol's color, but leave Alice's and Carol's pets unconstrained---both \{Alice:cat, Carol:fish\} and \{Alice:fish, Carol:cat\} are valid. \textbf{Fix:} brute-force all valid solutions (36~permutations) and accept any match. Fix the puzzle generator to produce uniquely solvable puzzles.

\paragraph{Class 6: R2+N2 word-number interaction (126 cases, 2.2\%).}
R2~(transitive ordering) requires a numbered list (``1.~Mouse, 2.~Rabbit...''). When N2~(no digits) co-occurs, models write ``one.~Mouse, two.~Rabbit...'' to satisfy both constraints. R2's parser cannot extract word-numbered lists---it looks for digit prefixes only. The model's ordering is correct, and it correctly avoids digits, but the verifier cannot recognize the format. \textbf{Fix:} add word-number parsing to R2's list extractor.

\paragraph{Class 7: F3/F4 format variants (124 cases, 2.2\%).}
Models use a different but functionally equivalent format. For F3~(bullet list), 101~models used asterisk bullets (\texttt{*~item}) instead of the required dash bullets (\texttt{-~item}). For F4~(code blocks), 23~models used indented code blocks (4~spaces) instead of fenced blocks (\texttt{```}). These are arguably genuine failures since the constraint explicitly specifies the format, but could also be considered format-variant compliance.

\paragraph{Class 8: R1 relation format mismatch (92 cases, 1.6\%).}
The probe prompt describes spatial relations as ``A is above B.'' Models faithfully reproduce ``is above'' in their JSON output. The verifier expects just ``above'' (without the ``is'' prefix). Exact string matching fails despite correct semantics. \textbf{Fix:} strip common prefixes (``is~'', ``are~'') before comparison; clarify expected format in the prompt.

\paragraph{Summary.}
The answer-tag mechanism (wrapping the final response in \texttt{<answer>} tags and scoring only tagged content) eliminates Classes~1 and~2, which together account for 4,613 discrepancies (80.5\% of all identified issues). The remaining 20\% require targeted verifier fixes. Ten of 30~constraints (S4, L2, L5, W3, F1, F2, M2, N1, and arguably F3, F4) have zero discrepancies---their verifiers are fully reliable.

\paragraph{CSE LLM-judge spot-check audit.}
To validate verifier accuracy on the expanded CSE benchmark (36~constraints, 15~models, 223{,}643~failed constraint checks), we conducted a systematic spot-check using Claude~Opus~4.6 as an independent LLM judge across nine parallel audit passes---one per constraint category (R2, S1, L1, N1/N3, W2/W4, O1/O2/O3, F1/F2/F4, R3, M1/M2/S6)---sampling 10--20 near-miss failures per pass (failures with continuous score~${\geq}0.80$, indicating the model nearly satisfied the constraint). For each sampled failure, the judge independently parsed the model's response text, re-evaluated the constraint, and compared its verdict against the stored verifier output; all flagged discrepancies were then manually verified by the authors to confirm or dismiss the finding, yielding four cross-cutting issues ranked by impact: (1)~the \texttt{extract\_answer()} function matching the \emph{first} rather than \emph{last} \texttt{<answer>} tag in responses where models quote the prompt instruction, affecting 7{,}581+ scored rows across all constraints; (2)~the \texttt{split\_sentences()} function treating JSON objects, markdown tables, and bullet lists as prose sentences, producing 33--59\% false failures among near-miss ordering violations (O1/O2/O3) and 64~false S6 verdicts; (3)~the R3 JSON parser spanning from the first~\texttt{\{} to last~\texttt{\}} across multi-object responses, causing 1{,}051 false failures (19.2\% of R3 failures; DeepSeek~V4~Pro CSR would increase from 15\% to 59\% with the fix); and (4)~five R2 list-parsing gaps (inline multi-item lines, parenthetical numbering, space-delimited formats) causing ${\sim}$27\% of sampled R2 failures to be false negatives where the model solved the ordering correctly but the parser could not extract it---Three of four issues were fixed and rescored before final analysis: (1)~\texttt{extract\_answer()} now uses \texttt{rfind()} to match the last structural \texttt{<answer>} tag, skipping quoted instruction text; (2)~a \texttt{strip\_non\_prose()} helper removes JSON blocks, markdown tables, and bullet lists before sentence splitting in the O1, O2, O3, S6, F4, and W4 verifiers; (3)~the R3 JSON parser now uses bracket-depth tracking to extract the first complete, schema-matching JSON object rather than spanning from first~\texttt{\{} to last~\texttt{\}}.
The R2 parser gaps (issue~4) remain as a known limitation; the detailed per-constraint audit with proposed code fixes is documented in the supplementary materials.

\section{Impossible Probe Experiment}
\label{app:impossible_probes}

Beyond measuring how performance \emph{degrades} under composition, CSE includes impossible probes---constraint combinations with a proof of unsatisfiability---to study how models \emph{prioritize} when compliance is structurally unachievable.

Every impossible probe requires a clean logical proof of unsatisfiability; mechanical verifier conflicts (e.g., code-fence parsing interference) are excluded because the model cannot reason about verifier internals.
This separates impossible probes (which test constraint understanding and prioritization) from merely difficult probes (which test capacity).

\subsection{Design}

The 444~impossible probes (19~types $\times$ 3~topic variants $\times$ up to 15~models $=$ 55{,}645 constraint checks) span three categories of impossibility.

\emph{Binary contradictions} pair two constraints that are logically incompatible regardless of parameters: forbidden letter vs.\ mandatory words containing that letter (L1${+}$L2), embedding digits vs.\ forbidding digits (N1${+}$N2), alternating vs.\ monotonic sentence lengths (O1${+}$O3), cross-sentence word bridge vs.\ all-unique words (R4${+}$W1).
Each has a clean proof: for example, R4 forces the last word of sentence~$i$ to equal the first word of sentence~$i{+}1$, so any response with ${\geq}2$ sentences must repeat at least one word, violating W1.

\emph{Parameter-induced contradictions} use constraint pairs that are compatible for most parameter values but impossible under specific choices enforced at probe generation time: a sentence count of 3 paired with an acrostic word of length 5 (S2${+}$L4), a forbidden letter that appears in the required hidden-message word (L1${+}$M1), or two constraints that independently fix sentence count to different values (S2${+}$S6).

\emph{Hierarchy and load tests} probe whether sacrifice priorities change under compositional pressure.
One probe pairs forbidden letter `r' with a transitive ordering puzzle whose items all contain `r' (\emph{rabbit, raccoon, raven, rooster, ram, robin})---there is no partial solution, so the model must choose which constraint to abandon entirely.
Another embeds the L1${+}$L2 contradiction among 2, 4, or 6 additional compatible filler constraints, testing whether the sacrifice ordering is stable as $k$ increases from 4 to 8.
A four-way probe (forbidden letter `e' $+$ mandatory words $+$ unique words $+$ exact word count) tests whether the strongest sacrifice priority erodes under multi-constraint tension.

\subsection{Results}
\label{app:impossible_results}

Across 55{,}645 constraint checks, 28.2\% of individual constraints were satisfied on impossible probes.
Zero probe${\times}$model pairs passed all constraints on any impossible probe---the deterministic verifiers correctly reject every impossible probe for every model.

\paragraph{Constraint survival ranking.}
\Cref{tab:sacrifice_full} reports per-constraint pass rates on impossible probes, ordered by survival rate.
Binary constraints that require no sustained tracking dominate the top (L5: 92.7\%, N2: 69.5\%, L2: 60.6\%), while precise-count constraints requiring sustained maintenance cluster at the bottom (S4: 1.1\%, S6: 2.9\%, M2: 3.0\%, S1: 3.2\%).

\begin{table}[h]
\caption{Per-constraint survival rate on 444~impossible probes.}
\label{tab:sacrifice_full}
\centering
\begin{scriptsize}
\begin{tabular}{lrrr@{\hskip 12pt}lrrr}
\toprule
\textbf{CID} & \textbf{Pass} & \textbf{Score} & $n$ & \textbf{CID} & \textbf{Pass} & \textbf{Score} & $n$ \\
\midrule
L5 & 92.7 & 100.0 & 2038 & O3 & 18.5 & 52.0 & 714 \\
N2 & 69.5 & 99.4 & 924 & N3 & 17.5 & 43.0 & 1363 \\
L6 & 61.1 & 83.5 & 1620 & O2 & 15.2 & 49.6 & 2632 \\
L2 & 60.6 & 63.0 & 1827 & M3 & 13.8 & 52.6 & 1547 \\
R1 & 53.3 & 53.6 & 1136 & N1 & 13.5 & 33.5 & 1043 \\
R3 & 48.5 & 51.8 & 1800 & O1 & 12.1 & 46.0 & 1072 \\
F2 & 46.5 & 49.5 & 1778 & O4 & 11.3 & 25.6 & 2077 \\
N4 & 45.2 & 51.6 & 1252 & W1 & 10.7 & 72.1 & 1156 \\
F1 & 43.6 & 48.9 & 2967 & S3 & 9.7 & 37.2 & 1008 \\
W3 & 43.2 & 74.2 & 1396 & R4 & 9.0 & 25.9 & 1051 \\
F3 & 41.3 & 46.0 & 641 & W5 & 8.9 & 23.8 & 1927 \\
F4 & 38.3 & 84.9 & 1501 & S5 & 6.7 & 28.7 & 2984 \\
W4 & 35.0 & 91.0 & 1365 & M1 & 6.2 & 24.5 & 1021 \\
L1 & 32.9 & 98.4 & 2514 & S1 & 3.2 & 46.6 & 1651 \\
S2 & 31.7 & 62.7 & 833 & M2 & 3.0 & 26.4 & 1766 \\
L4 & 28.5 & 48.4 & 2161 & S6 & 2.9 & 11.4 & 1637 \\
W2 & 25.0 & 94.5 & 1759 & S4 & 1.1 & 23.7 & 1706 \\
R2 & 21.3 & 37.4 & 550 \\
\bottomrule
\end{tabular}
\end{scriptsize}
\end{table}

\paragraph{Sacrifice rules.}
Three deterministic rules emerge across all 15~models:
\begin{enumerate}[leftmargin=*, itemsep=2pt]
\item \textbf{Concrete inclusion ${>}$ abstract avoidance.} L2~(mandatory words) survives at 73--100\%; L1~(lipogram) is sacrificed at 0--7\%. Models universally keep the named word and accept the letter violation.
\item \textbf{Prohibition ${>}$ requirement.} N2~(no digits) survives at 84--91\%; N1 and N3 are sacrificed at 0\%, N4 at 4\%. Models find it easier to omit than to include.
\item \textbf{Natural structure ${>}$ imposed structure.} O3~(alternating) survives at 56\%; O1~(monotonic) at 7\% and O4~(growth) at 0\%. Alternating sentence lengths approximate natural prose more closely than strict monotonic or Fibonacci growth.
\end{enumerate}

\paragraph{Hierarchy inversion under binary choice.}
When forbidden letter `r' is paired with a transitive ordering puzzle whose items all contain `r', the model must abandon one constraint entirely.
R2~(relational, the fastest degrader under compositional load) survived at 33\%, while L1~(lexical, compositionally immune) was sacrificed at 7\%---the opposite of the degradation hierarchy.
The same inversion appears when cross-sentence bridge (R4) is paired with unique words (W1): R4 survives at 62\%, W1 is sacrificed at 2\%.
Both cases are suggestive of a sunk-cost mechanism: the model commits computation to the harder constraint (solving the ordering puzzle, maintaining the bridge pattern) and preserves it over the simpler one (letter avoidance, word uniqueness).
These results rest on $N{=}3$ probes each and require replication.

\paragraph{Stability under compositional load.}
When the L1${+}$L2 contradiction is embedded among 2--6 additional compatible constraints ($k{=}4$ to $k{=}8$), the sacrifice ordering is stable: L2 survives at 82--89\% and L1 is sacrificed at 0--4\% regardless of $k$.
In a four-way probe (forbidden letter `e' $+$ mandatory words $+$ unique words $+$ exact word count), L2 survives at 100\% and L1 at 0\%---the strongest sacrifice priority is unanimous even under multi-constraint tension.
Compositional load compresses sacrifice margins slightly but does not alter the ordering.

\paragraph{Budget-exhaustion regime.}
On the 389~impossible probes where the model must triage across many jointly unsatisfiable constraints---rather than choosing between two that directly contradict---a second sacrifice regime emerges that \emph{preserves} the degradation hierarchy rather than inverting it.
Binary constraints that require no sustained tracking survive at the highest rates (L5: 90--100\%, N2: 56--83\%, L2: 53--73\%).
One-shot structural constraints also survive (F1: 41--60\%, R1: 50--60\%): format and relational JSON output require a single planning decision rather than sustained maintenance.
The consistently sacrificed constraints are precisely those with the highest sustained-tracking demand (S4: 0--7\%, S6: 0--6\%, M2: 0--6\%, S1: 2--7\%), matching the degradation hierarchy from \cref{tab:constraint_profiles}.

\paragraph{Two regimes of sacrifice.}
The sacrifice ordering depends on the choice structure.
Under \emph{binary choice}---where one of two constraints must go---models preserve whichever constraint required greater planning investment, inverting the degradation hierarchy.
Under \emph{multi-constraint triage}---where the model must decide what it can fit from a jointly unsatisfiable set---the sacrifice ordering preserves the degradation hierarchy: the hardest constraints are dropped first, the easiest survive.

% COMMENTED OUT — table spills column, Claude 4.7 row is stale (pre-16K)
% \section{Answer Tag Compliance}
% \label{app:answer_tags}
% [Full content preserved — restore after fixing Claude 4.7 row and table width]

\section{Comprehension vs.\ Execution: A Case Study}
\label{app:comp_vs_exec}

\begin{figure*}[t]
\centering
\includegraphics[width=\textwidth]{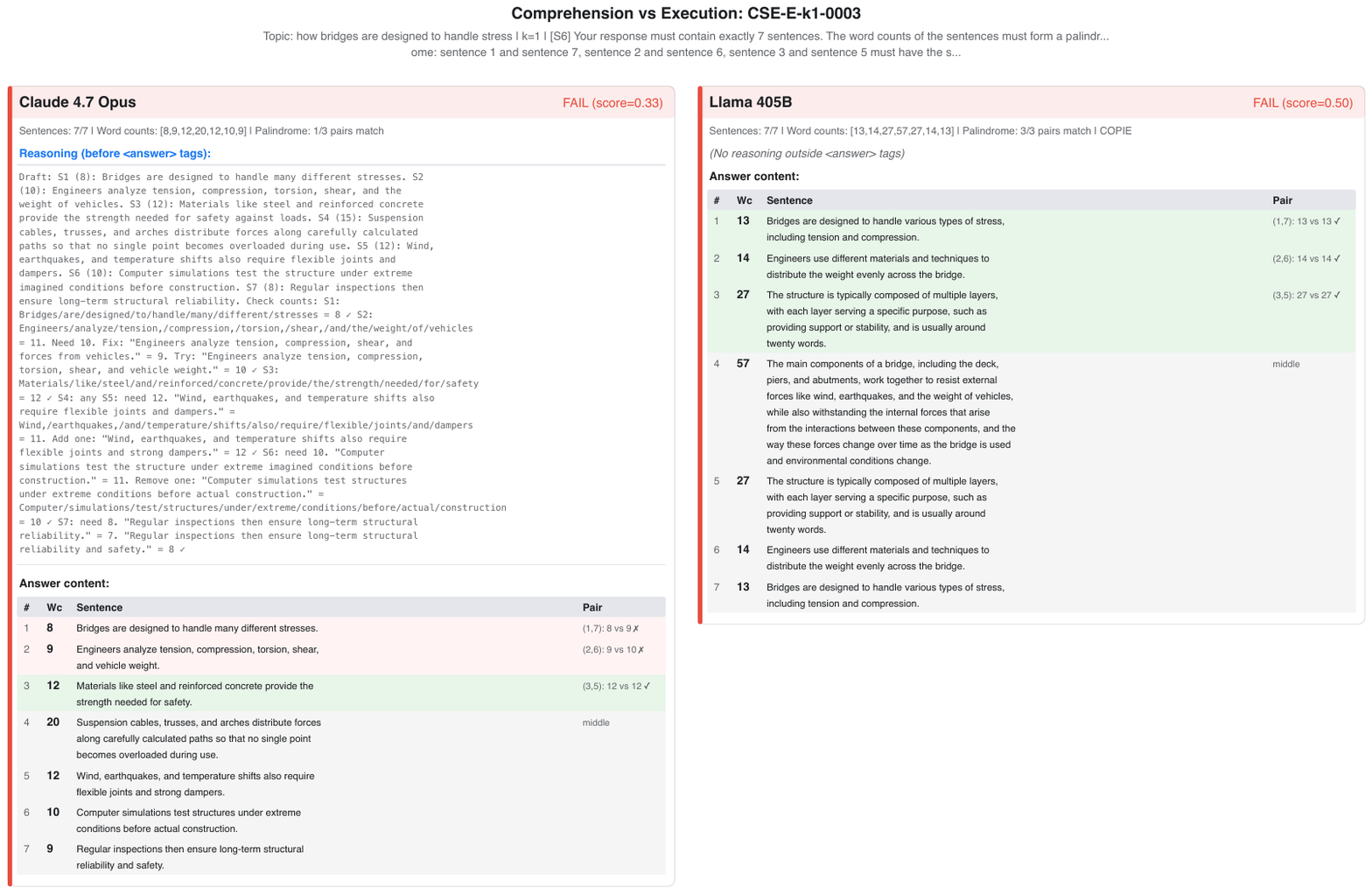}
\caption{Comprehension vs.\ execution on probe CSE-E-k1-0003 (S6: palindromic sentence word counts, $k{=}1$). \textbf{Left:} Claude~4.7~Opus drafts, counts, and revises---demonstrating full comprehension of the palindrome structure---but miscounts by one word on 2/3 pairs (score~0.33). \textbf{Right:} Llama~405B copies sentences~1--3 as~5--7 reversed, mechanically guaranteeing a perfect palindrome (score~1.0) without explicit reasoning. The constraint requires palindromic word counts, not unique content; Llama's copy strategy is a legitimate exploitation of this specification gap. Models that find mechanical shortcuts to satisfy constraints outperform models that reason about them explicitly---the comprehension-maintenance gap is about strategy selection, not just tracking difficulty.}
\label{fig:comp_vs_exec}
\end{figure*}

\Cref{fig:comp_vs_exec} presents a side-by-side comparison of two models responding to the same probe (CSE-E-k1-0003, constraint S6: palindromic sentence structure). The probe requires exactly 7~sentences whose word counts form a palindrome---sentences~1 and~7, 2~and~6, 3~and~5 must have the same word count.

Claude~4.7~Opus reasons about the constraint explicitly: it drafts sentences, counts words, catches errors (``S2: 11.\ Need 10.\ Fix\ldots''), revises, and verifies. Its chain-of-thought trace shows genuine comprehension of the palindromic structure. Yet the final answer has word counts [8, 9, 12, 20, 12, 10, 9]---pairs (1,7) are 8~vs.~9, (2,6) are 9~vs.~10. Off by one word on two of three pairs (score~=~0.33).

Llama~405B bypasses reasoning entirely. It copies sentences~1--3 as sentences~5--7 in reverse order, producing word counts [13, 14, 27, 57, 27, 14, 13]---a perfect palindrome (score~=~1.0). The sentences are not merely matched in word count; they are \emph{identical}, which mechanically guarantees equal counts.

This is not a violation of the constraint. The specification requires palindromic \emph{word counts}, not unique sentence content. Llama's strategy exploits a specification gap: copying is the most reliable way to guarantee exact word-count matches. Preventing this would require an additional constraint (``all sentences must be distinct''), which would itself interact with W4 (no repeated bigrams) and significantly alter the compositional landscape.

This case study illustrates that the comprehension-maintenance gap is not solely about tracking difficulty---it is also about \emph{strategy selection}. Models that find mechanical shortcuts to satisfy constraints outperform models that reason about them explicitly. Claude~4.7's failure is not one of understanding (it comprehends the palindrome structure perfectly) but of execution (it cannot maintain exact word counts across revisions). The ``dumber'' strategy wins precisely because it eliminates the execution bottleneck entirely.

\end{document}